\documentclass{article}

\PassOptionsToPackage{numbers, compress}{natbib}

\usepackage[preprint]{neurips_2023}

\usepackage[utf8]{inputenc} 
\usepackage[T1]{fontenc}    
\usepackage{url}            
\usepackage{booktabs}       
\usepackage{amsfonts}       
\usepackage{nicefrac}       
\usepackage{microtype}      
\usepackage{xcolor}         

\usepackage{graphicx}
\usepackage{amsmath}
\usepackage{amssymb}
\usepackage{makecell}

\usepackage{multirow}
\usepackage{multicol}
\usepackage{colortbl} 
\definecolor{blush}{rgb}{0.87, 0.36, 0.51}
\definecolor{Gray}{gray}{0.7}
\newcolumntype{a}{>{\columncolor{Gray}}r}
\definecolor{DimGray}{gray}{0.9}
\newcolumntype{d}{>{\columncolor{DimGray}}r}
\newcommand{\cmark}{\ding{51}}%
\newcommand{\xmark}{\ding{55}}%

\usepackage{pifont}

\usepackage{tabularx}
\usepackage{listings}
\usepackage{algorithm,algpseudocode}
\usepackage[pagebackref,breaklinks,colorlinks]{hyperref}
\usepackage{amsmath}
\usepackage{dsfont}
\usepackage{caption}

\title{PseudoCal: A Source-Free Approach to Unsupervised Uncertainty Calibration in Domain Adaptation}

\author{%
  Dapeng Hu$^{1,2}$ ~~~ Jian Liang$^{4}$ ~~~ Xinchao Wang$^5$ ~~~ Chuan-Sheng Foo$^{3,1}$~\thanks{Corresponding author} \\
  $^1$Centre for Frontier AI Research, A*STAR, Singapore\\
  $^2$Institute of High Performance Computing, A*STAR, Singapore\\
  $^3$Institute for Infocomm Research, A*STAR, Singapore\\
  $^4$CRIPAC $\&$ MAIS, Institute of Automation, Chinese Academy of Sciences\\
  $^5$National University of Singapore\\
  \texttt{lhxxhb15@gmail.com, liangjian92@gmail.com, }\\
  \texttt{xinchao@nus.edu.sg, foo$\_$chuan$\_$sheng@i2r.a-star.edu.sg} 
}

\begin{document}

\maketitle

\begin{abstract}
 Unsupervised domain adaptation (UDA) has witnessed remarkable advancements in improving the accuracy of models for unlabeled target domains. However, the calibration of predictive uncertainty in the target domain, a crucial aspect of the safe deployment of UDA models, has received limited attention. The conventional in-domain calibration method, \textit{temperature scaling} (TempScal), encounters challenges due to domain distribution shifts and the absence of labeled target domain data. Recent approaches have employed importance-weighting techniques to estimate the target-optimal temperature based on re-weighted labeled source data. Nonetheless, these methods require source data and suffer from unreliable density estimates under severe domain shifts, rendering them unsuitable for source-free UDA settings.
To overcome these limitations, we propose PseudoCal, a source-free calibration method that exclusively relies on unlabeled target data. Unlike previous approaches that treat UDA calibration as a \textit{covariate shift} problem, we consider it as an unsupervised calibration problem specific to the target domain. Motivated by the factorization of the negative log-likelihood (NLL) objective in TempScal, we generate a labeled pseudo-target set that captures the structure of the real target. By doing so, we transform the unsupervised calibration problem into a supervised one, enabling us to effectively address it using widely-used in-domain methods like TempScal.
Finally, we thoroughly evaluate the calibration performance of PseudoCal by conducting extensive experiments on 10 UDA methods, considering both traditional UDA settings and recent source-free UDA scenarios. The experimental results consistently demonstrate the superior performance of PseudoCal, exhibiting significantly reduced calibration error compared to existing calibration methods.
\end{abstract}
\section{Introduction}
\label{sec:intro}
In recent years, unsupervised domain adaptation (UDA)\cite{pan2009survey, pan2010domain} has become a popular technique for effectively improving the generalization of deep learning models\cite{alexnet12, resnet16, dosovitskiy2020image} from labeled source datasets to unlabeled out-of-domain target datasets.
Remarkable strides have been made in the development of novel UDA methods~\cite{ganin2015unsupervised, sun2016deep, tzeng2017adversarial, long2018conditional, saito2018maximum, liang2021domain}, practical UDA applications~\cite{chen2018domain, saito2019strong, tsai2018learning, vu2019advent}, and real-world UDA settings~\cite{long2015learning, cao2018partial, panareda2017open, liang2020we}. 
Despite such advancements in UDA, there is a predominant focus on improving the performance of deep learning models in the target domain, while the calibration of target predictive uncertainty remains largely unexplored. This aspect is crucial for the deployment of UDA models in safety-critical decision-making scenarios, as deep learning models are known to suffer from the miscalibration problem, where confidence does not accurately reflect the likelihood of correctness~\cite{guo2017calibration, lakshminarayanan2017simple}.
Recent seminal works~\cite{park2020calibrated, pampari2020unsupervised, wang2020transferable} have notably addressed the challenge of uncertainty calibration in UDA by focusing on the assumption of \textit{covariate shift}~\cite{sugiyama2007covariate}.
They commonly employ \textit{importance weighting}~\cite{cortes2008sample} to re-weight the labeled source validation data for target-adaptive \textit{temperature scaling} (TempScal)~\cite{guo2017calibration}. However, these approaches have certain limitations that need to be addressed. Firstly, \textit{importance weighting} is not reliable for large \textit{covariate shift} and label shift scenarios~\cite{park2020calibrated}. Secondly, these methods require access to source data, making them unsuitable for privacy-preserving source-free UDA settings~\cite{li2020model, liang2020we, liang2022dine}. Lastly, the additional model training and density estimation involved in these methods make them more complex compared to the simple and post-hoc method of TempScal.

\begin{figure}[!htbp]
\centering
\footnotesize
\setlength\tabcolsep{2.0 mm}
\renewcommand\arraystretch{0.05}
\begin{tabular}{cc}
\includegraphics[width=0.48\linewidth,trim={0.0cm 0.0cm 0.0cm 0.0cm}, clip]{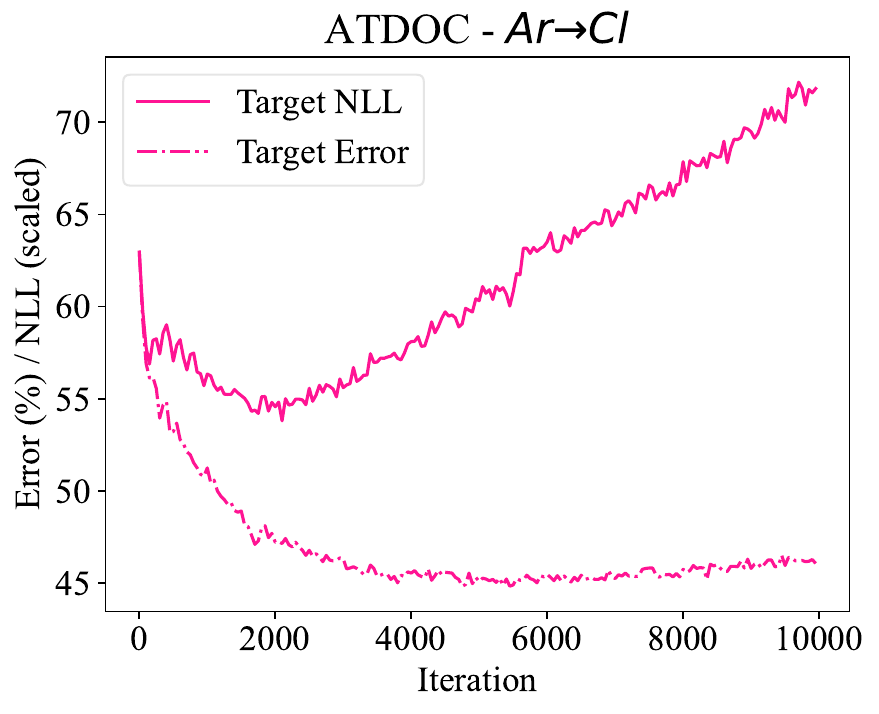} &
\includegraphics[width=0.48\linewidth,trim={0.0cm 0.0cm 0.0cm 0.0cm}, clip]{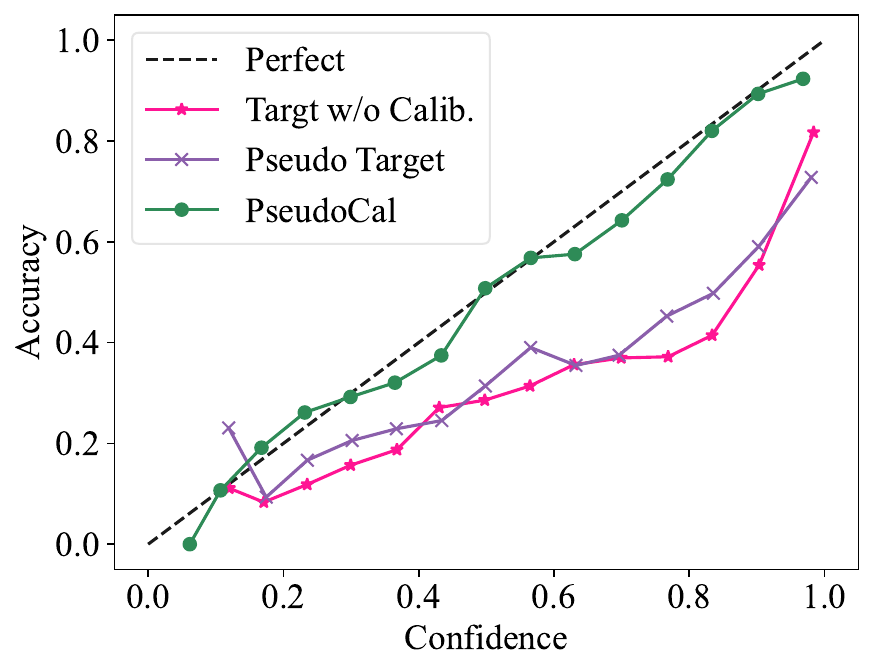} \\
~\\
(a) Overfitting of NLL in UDA training & (b) Reliability diagrams  \\
\end{tabular}
\caption{ATDOC~\cite{liang2021domain} on a closed-set UDA task Ar $\to$ Cl. (a) illustrates the target error and target negative log-likelihood (NLL, rescaled to match error) during UDA training. (b) shows reliability diagrams~\cite{guo2017calibration} for the real target and pseudo target. Perfect: the ideal case without any miscalibrations.}
\label{fig:atdoc-fig}
\end{figure}

To address these limitations, this paper aims to tackle the challenge of predictive uncertainty calibration in the unlabeled target domain without relying on source data. Unlike existing approaches that treat uncertainty calibration in UDA as a \textit{covariate shift} problem, we adopt a distinct perspective by considering it as \textit{an unsupervised calibration problem in the target domain}. Inspired by the pioneering work of Guo \textit{et al.}~\cite{guo2017calibration}, we compare the target error and scaled target negative log-likelihood (NLL) in Figure~\ref{fig:atdoc-fig} (a). The figure clearly shows that the target NLL encounters significant overfitting during training of UDA, which aligns with similar observations in learning scenarios involving independent and identically distributed (IID) data~\cite{guo2017calibration}. Moreover, by factorizing the NLL objective employed in IID TempScal, we uncover that both correct and wrong predictions contribute to the final optimized temperature. As a result, we put forth the hypothesis that \textit{the target-domain oracle temperature can be accurately approximated by optimizing TempScal using data that share a similar accuracy-uncertainty distribution as real target data}.

Based on this hypothesis, we introduce our source-free approach called pseudo-target calibration (PseudoCal). PseudoCal begins by synthesizing a `labeled' dataset comprising pseudo-target samples and corresponding pseudo-target labels, generated using \textit{mixup}~\cite{zhang2018mixup} with real target samples and pseudo labels. Remarkably, we observe that the pseudo-target set exhibits a similar accuracy-confidence distribution to the real target set, as demonstrated in Figure~\ref{fig:atdoc-fig} (b). Such similarity can be attributed to the well-known ~\textit{cluster assumption}~\cite{grandvalet2004semi, chapelle2005semi, lee2013pseudo, shu2018dirt, verma2022interpolation}, where samples located far away from the decision boundary are more likely to be correctly classified, while those near the decision boundary are prone to misclassification. Building on this assumption, we can establish correspondences between the pseudo-target set and real target set, where correctly predicted pseudo-target samples correspond to high-margin real samples, and wrongly predicted pseudo-target samples correspond to low-margin real samples. Such correspondences easily convert the unsupervised calibration problem into a supervised one. Consequently, our PseudoCal can estimate the \textit{Oracle} real temperature by utilizing the pseudo temperature obtained through supervised TempScal optimization on the pseudo-target set.

We make three primary contributions in this paper.
\begin{itemize}
 \item We address the understudied challenge of predictive uncertainty calibration in unsupervised domain adaptation (UDA) from a novel source-free perspective. Unlike existing approaches that treat UDA calibration as a \textit{covariate shift} problem, we consider it as an unsupervised calibration problem in the target domain. This unique perspective unifies calibration in UDA across different settings, including scenarios with label shift or limited source access.
 \item We introduce a novel source-free and post-hoc approach, namely pseudo-target calibration (PseudoCal), for UDA calibration. By leveraging the \textit{cluster assumption}, PseudoCal successfully converts the unsupervised calibration problem into a more manageable supervised problem. PseudoCal achieves this by generating a `labeled' pseudo-target set through \textit{mixup} and employing supervised TempScal optimization on this dataset to estimate the pseudo temperature used for the real target samples.
 
 \item We conduct a comprehensive evaluation of PseudoCal and compare it with 7 existing calibration baselines in UDA. Specifically, we conduct experiments on 10 UDA methods across 5 challenging UDA scenarios, spanning diverse UDA benchmarks including both image classification and segmentation tasks. The calibration results consistently demonstrate that, on average, PseudoCal significantly outperforms all of the competing methods.
\end{itemize}

\section{Related Work}
\label{sec:relatedwork}

\textbf{Unsupervised domain adaptation.} Unsupervised domain adaptation (UDA) has witnessed notable progress, evident in the proposal of various effective UDA approaches, the extension to diverse machine learning tasks, and the exploration of a wide range of real-world settings. UDA has been extensively studied in image classification tasks, where existing state-of-the-art methods can be categorized into two main lines: (1) distribution alignment across domains using specific discrepancy measures~\cite{long2015learning, sun2016deep} or adversarial learning~\cite{ganin2015unsupervised, tzeng2017adversarial, long2018conditional, saito2018maximum}, and (2) target domain-based learning with self-training~\cite{shu2018dirt, liang2021domain} or regularizations~\cite{xu2019larger, cui2020towards, jin2020minimum}.
Moreover, UDA has also been studied in object detection~\cite{chen2018domain, saito2018maximum} and image segmentation~\cite{tsai2018learning, vu2019advent}. 
Initially, UDA is based on the \textit{covariate shift} assumption~\cite{sugiyama2007covariate}, which means that two domains share similar label and conditional distributions but have different input distributions. This is commonly referred to as closed-set UDA. In recent years, several new practical settings have emerged to address additional challenges. These settings further consider label shift~\cite{lipton2018detecting}, including partial-set UDA~\cite{cao2018partial,liang2020balanced}, where some source classes are missing in the target domain, and open-set UDA~\cite{panareda2017open}, where the target domain contains samples from unknown classes. 
Recently, there has been a growing interest in a novel practical setting called source-free UDA, which focuses on preserving source privacy. Source-free UDA encompasses two main settings: the white-box setting~\cite{li2020model, liang2020we}, where the source model is available for target adaptation, and the more stringent black-box setting~\cite{zhang2021unsupervised, liang2022dine}, where the source model is solely utilized for inference purposes.

\textbf{Uncertainty calibration.}
The study of uncertainty calibration begins with techniques such as \textit{histogram binning}~\cite{zadrozny2001obtaining}, \textit{isotonic regression}~\cite{zadrozny2002transforming}, and \textit{Platt scaling}~\cite{platt1999probabilistic}, initially applied to binary classification tasks. Guo \textit{et al.}\cite{guo2017calibration} extends \textit{Platt scaling} to multi-class classification and introduces \textit{matrix scaling} (MatrixScal), \textit{vector scaling}(VectorScal), and \textit{temperature scaling} (TempScal). These post-hoc methods require a labeled validation set for calibration. On the other hand, there are methods that address calibration during model training, including \textit{Monte Carlo Dropout} (MC-Dropout)\cite{gal2016dropout}, Ensemble~\cite{lakshminarayanan2017simple}, and \textit{Stochastic Variational Bayesian Inference} 
(SVI)~\cite{blundell2015weight, louizos2017multiplicative, wen2018flipout}. However, an evaluation in~\cite{ovadia2019can} reveals that these methods do not maintain calibration performance under dataset shift.
In addition to calibration in IID settings and classification tasks, there is growing interest in calibration under distribution shifts~\cite{alexandari2020maximum, wang2020transferable, park2020calibrated} and in semantic segmentation tasks~\cite{ding2021local, wang2022calibrating, de2023reliability}. In this paper, we specifically address the calibration problem in single-source unsupervised domain adaptation (UDA).
Various calibration methods have been proposed to handle domain distribution shifts. The first type utilizes \textit{importance weighting}~\cite{cortes2008sample} to address calibration under \textit{covariate shift} in UDA, exemplified by CPCS~\cite{park2020calibrated} and TransCal~\cite{wang2020transferable}. The second type involves perturbing the source validation set to serve as a general target set~\cite{tomani2021post, salvador2021improved}. More recently, some methods~\cite{gong2021confidence, yu2022robust} have utilized multiple source domains to calibrate the unlabeled target domain in UDA. Additionally, there are training-stage calibration methods that employ label smoothing~\cite{thulasidasan2019mixup, liu2022devil} or optimize accuracy-uncertainty differentiably~\cite{krishnan2020improving}.
Among these methods, CPCS and TransCal are noteworthy as they specifically address transductive target calibration in UDA.
For more general approaches like MC-Dropout and Ensemble, we compare our method directly with Ensemble because it consistently outperforms MC-Dropout.
Table~\ref{tab:method-cmp} presents a comprehensive comparison of these typical UDA calibration methods. Our proposed method, PseudoCal, distinguishes itself through its simplicity, achieving source-free calibration with a single source model.

\begin{table*}[!htbp]\centering
\caption{Comparisons of typical calibration methods in unsupervised domain adaptation.}
\resizebox{0.92\textwidth}{!}{$
\begin{tabular}{l|cc|ccc}
\toprule[1pt]
Calibration Method & covariate shift & label shift & no harm on accuracy & no extra training & no source data \\
\midrule
TempScal~\cite{guo2017calibration} & \xmark & \xmark & \cmark & \cmark & \xmark \\
MC-Dropout~\cite{gal2016dropout} & \cmark & \cmark & \xmark & \cmark & \cmark \\
Ensemble~\cite{lakshminarayanan2017simple} & \cmark  & \cmark & \cmark & \xmark & \cmark \\
CPCS~\cite{park2020calibrated}  & \cmark& \xmark & \cmark & \xmark & \xmark \\
TransCal~\cite{wang2020transferable}   & \cmark& \xmark & \cmark & \xmark & \xmark \\
PseudoCal (Ours)   & \cmark & \cmark & \cmark & \cmark & \cmark \\
\bottomrule[1pt]
\end{tabular}%
$}
\label{tab:method-cmp}%
\end{table*}%

\section{Approach}
\label{sec:method}
In this paper, we address the problem of predictive uncertainty calibration in the context of unsupervised domain adaptation (UDA).
We begin by introducing UDA with a $C$-way image classification task. UDA involves two domains: a labeled source domain and an unlabeled target domain.
The source domain $\mathcal{D}_{\rm{s}}=\{(x_{\rm{s}}^{i}, y_{\rm{s}}^{i})\}_{{i=1}}^{n_{\rm{s}}}$ consists of $n_{\rm{s}}$ images $x_{\rm{s}}$ with their corresponding labels $y_{\rm{s}}$, where $x_{\rm{s}}^i \in \mathcal{X}_{\rm{s}}$ and $y_{\rm{s}}^i \in \mathcal{Y}_{\rm{s}}$.
The target domain $\mathcal{D}_{\rm{t}}=\{x_{\rm{t}}^{i}\}_{{i=1}}^{n_{\rm{t}}}$ contains unlabeled images $x_{\rm{t}}$, where $x_{\rm{t}}^i \in \mathcal{X}_{\rm{t}}$.
The objective of UDA is to learn a UDA model $\phi$ that can predict the unknown ground truth labels $\{y_{\rm{t}}^{i}\}_{{i=1}}^{n_{\rm{t}}}$ for the target domain, utilizing data from both domains simultaneously~\cite{ganin2015unsupervised} or sequentially~\cite{liang2020we}. In addition to \textit{covariate shift}, we also tackle label shift in partial-set UDA \cite{liang2020balanced}.

\subsection{Calibration Metrics}

Next, we introduce the calibration problem and relevant metrics. When feeding a random sample $(x, y)$ into the UDA model $\phi$, we can obtain the predicted class $\hat{y}$ and the corresponding softmax-based confidence $\hat{p}$. Ideally, the confidence should accurately reflect the probability of correctness, expressed as $\mathbb{P} (\hat{y}=y | \hat{p}=p) = p,\ \forall \ p \in [0, 1]$. This perfect calibration, also known as \textit{Perfect}, is impossible to achieve~\cite{guo2017calibration}. The widely used metric for evaluating calibration error is the expected calibration error (ECE)~\cite{guo2017calibration}. ECE involves partitioning probability predictions into $M$ bins, with $B_m$ representing the indices of samples falling into the $m$-th bin. It calculates the weighted average of the accuracy-confidence difference across all bins:
\vspace{-5pt}
\begin{equation*}
\label{ece}
\begin{aligned}
\mathcal{L}_{\rm{ECE}} =  \sum_{m=1}^M \frac {|B_m| } {n} | \rm{acc} ( B_m )  - \rm{conf} ( B_m )|
\end{aligned}
\end{equation*}
Here, $n$ represents the number of samples, and for the $m$-th bin, the accuracy is computed as $\rm{acc}$ $(B_m) = $ ${|B_m|^{-1} } \sum_{i \in B_m} \mathds{1}(\hat{y}_i = y_i)$, and the confidence is computed as $\rm{conf}$ $(B_m) = $ ${|B_m|^{-1} } \sum_{i \in B_m} \hat{p_i}$. The introduction of additional popular metrics, such as NLL and Brier Score (BS)~\cite{brier1950verification}, is provided in the appendix for further reference.

\subsection{Factorized Temperature Scaling}
\textit{Temperature scaling} (TempScal)~\cite{guo2017calibration} is a widely employed calibration method in IID learning scenarios due to its simplicity and effectiveness. It is a post-hoc calibration technique that optimizes a temperature scalar, denoted as $T$, on a labeled validation set using the negative log-likelihood (NLL) loss between the temperature-flattened softmax predictions and the ground truth labels.
For the unlabeled target domain in UDA, we define the calibration achieved by applying TempScal with raw predictions and unattainable target ground truths as the `Oracle' calibration. This serves as an upper bound for all other calibration methods.
Let $z$ represent the corresponding logit vector for the image input $x$, and let $\sigma(\cdot)$ denote the softmax function. The `Oracle' target temperature, denoted as $T_{\rm{o}}$, can be obtained using the original \textit{temperature scaling} optimization formulated as follows 
\vspace{-3pt}
\begin{equation}
\label{tempscal}
\begin{aligned}
	T_{\rm{o}} =  \mathop {\arg \min }\limits_{T} \, {\mathbb{E}_{(x_i, y_i) \in \mathcal{D}_{\rm{t}}}} \ {\mathcal{L}_{{\rm{NLL}}}}\left({\sigma (z_i/ T)}, y_i \right) \\
 \end{aligned}
\end{equation}
Upon closer examination of TempScal, we observe that samples in the validation set can be classified as either correctly or wrongly predicted. Further, both types of samples have contrasting effects on the temperature optimization process. Specifically, the NLL minimization favors a small temperature to sharpen the confidence with correct predictions and a large temperature to flatten the confidence with wrong predictions. As a result, we can decompose Equation~\ref{tempscal} as follows:
\begin{equation}
\label{fact_tempscal}
\begin{aligned}
	T_{\rm{o}} &= \mathop {\arg \min }\limits_{T} \, 
        \frac{N_{\rm{c}}}{N} {\mathbb{E}_{(x_i, y_i) \in \mathcal{D}_{\rm{c}}}} \ {\mathcal{L}_{{\rm{NLL}}}}\left( {\sigma (z_i/ T)}, y_i \right) + \frac{N_{\rm{w}}}{N} {\mathbb{E}_{(x_j, y_j) \in \mathcal{D}_{\rm{w}}}} \ {\mathcal{L}_{{\rm{NLL}}}}\left( {\sigma (z_j/ T)}, y_j \right), \\
 \end{aligned}
\end{equation}
where $\mathcal{D}_{\rm{c}}$ represents the dataset of correctly predicted samples, comprising $N_{\rm{c}}$ instances. Similarly, $\mathcal{D}_{\rm{w}}$ denotes the dataset of wrongly predicted samples, consisting of $N_{\rm{w}}$ instances.

\begin{figure}[!t]
\centering
\footnotesize
\setlength\tabcolsep{1.0mm}
\renewcommand\arraystretch{0.1}
\begin{tabular}{c}
\includegraphics[width=0.88\linewidth,trim={0.0cm 0.0cm 0.0cm 0.0cm}, clip]{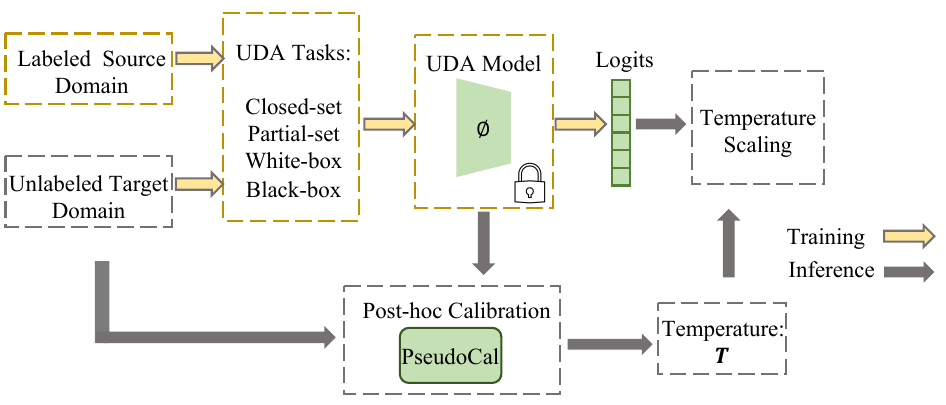} \\
\end{tabular}
\caption{The pipeline of PseudoCal for uncertainty calibration in UDA.}
\label{fig:pipeline}
\end{figure}

\subsection{PseudoCal: Pseudo-Target Calibration}

\textbf{Motivation.}
We propose an innovative perspective on uncertainty calibration in UDA by reframing it as an unsupervised calibration problem in the target domain, completely independent of source data. Examining the factorization in Equation~\ref{fact_tempscal}, we observe that if two data sets exhibit a similar correct-wrong pattern, they should also share a similar temperature when using TempScal. This observation motivates our hypothesis: if we can synthesize a labeled pseudo-target set with a similar correct-wrong pattern as the real target set, we can obtain a reliable estimation of the target oracle temperature even without applying TempScal directly to the real target.

However, modeling the correct-wrong pattern of the real target directly is infeasible without target labels. The presence of domain shift often leads to significant deviations between predicted pseudo-labels and ground truth labels, rendering calibration with raw predictions and pseudo-labels unreliable. This is demonstrated in our experiments (Table~\ref{tab:abla-mix}). To address this issue, we propose synthesizing samples to approximate the accuracy-confidence distribution of the real target.
In contrast to other augmentation techniques involving random perturbations~\cite{cubuk2020randaugment, chen2020improved} or vicinal perturbations~\cite{miyato2018virtual}, we find that \textit{mixup} provides a simple approach to generate controlled cross-cluster perturbations. Notably, the mixed samples naturally encompass both correct and wrong predictions, aligning with the cluster assumption~\cite{verma2022interpolation} that we will discuss later in our analysis.

\textbf{Pseudo-target synthesis via \textit{mixup}.}
We first generate a pseudo-target set by applying the \textit{mixup} technique~\cite{zhang2018mixup} to all target samples. Specifically, a pseudo-target sample $x_{\rm{pt}}$ and its label $y_{\rm{pt}}$ are obtained by taking a convex combination of a pair of real target samples ${x_{\rm{t}}^i, x_{\rm{t}}^j}$ and the different predicted pseudo labels ${\hat{y}_{\rm{t}}^i, \hat{y}_{\rm{t}}^j}$. Consequently, we obtain a labeled pseudo-target set $\{(x_{\rm{pt}}^{i}, y_{\rm{pt}}^{i})\}_{{i=1}}^{n_{\rm{pt}}}$, where $n_{\rm{pt}}$ represents the amount. The process of pseudo-target synthesis is formulated as follows:
\vspace{-3pt}
\begin{equation}
\label{eq:mixup}
x_{\rm{pt}} = \lambda * x_{\rm{t}}^{i} + (1 - \lambda) * x_{\rm{t}}^{j}, \qquad y_{\rm{pt}} = \lambda * \hat{y}_{\rm{t}}^{i} + (1 - \lambda) * \hat{y}_{\rm{t}}^{j},
\end{equation}
where $\lambda$ is a fixed scalar used as the mix ratio.

\textbf{Supervised calibration with \textit{temperature scaling}.}
Using the generated labeled pseudo-target set $\{(x_{\rm{pt}}^{i}, y_{\rm{pt}}^{i})\}_{{i=1}}^{n_{\rm{pt}}}$, we can easily determine the optimal pseudo-target temperature through supervised methods such as TempScal. This estimated temperature serves as an approximation of the `Oracle' target temperature. With this step, we effectively transform the challenging unsupervised calibration problem associated with the real target set into a supervised one using the pseudo-target set. The source-free calibration pipeline of PseudoCal is illustrated in Figure~\ref{fig:pipeline}, where the UDA model is utilized as a black box solely for inference. 
We compare the accuracy-confidence distribution between the real target and pseudo target, and present the calibration performance of PseudoCal in comparison to the vanilla no calibration case in Figure~\ref{fig:atdoc-fig} (b), providing strong evidence to support our hypothesis and validate the effectiveness of PseudoCal.

\textbf{Analysis through the lens of the \textit{cluster assumption}.}
We offer an intuitive analysis of why \textit{mixup} facilitates the synthesis of a pseudo-target set with a similar accuracy-confidence distribution to the real target.
Our analysis is grounded in the widely accepted and theoretically justified \textit{cluster assumption}~\cite{grandvalet2004semi, chapelle2005semi}, which has been extensively applied in semi-supervised learning~\cite{lee2013pseudo, laine2016temporal, tarvainen2017mean,sohn2020fixmatch, verma2022interpolation} and domain adaptation~\cite{morerio2017minimal, shu2018dirt, mishra2021surprisingly}.
According to the \textit{cluster assumption}, the decision boundary should reside in low-density regions of a learned cluster structure. This implies that samples located far from the decision boundary are more likely to be correctly classified, whereas those near the boundary are prone to misclassification.

In a UDA task, the model $\phi$ is typically well-trained to learn a target structure. When employing \textit{mixup}, the hard pseudo-target label of a pseudo-target sample is determined by the dominant real sample with a mix ratio exceeding $0.5$.
Consequently, we can expect the following correspondence in terms of the accuracy-confidence distribution between the real target and pseudo target: (\textit{i}) pseudo-target samples with correct predictions matching pseudo-target labels indicate that their dominant real samples also possess correct real predictions, (\textit{ii}) conversely, pseudo-target samples with wrong predictions mismatching pseudo-target labels indicate that their dominant real samples have wrong real predictions.
This correspondence provides a certain degree of guarantee for the success of PseudoCal.
We empirically demonstrate the robustness of such a guarantee across various UDA tasks. Remarkably, even when applied to a weak UDA model with a target accuracy of only $30\%$, PseudoCal consistently exhibits substantial improvements in calibration.

\section{Experiments}
\label{sec:experiment}

\subsection{Settings}
\textbf{Datasets.}
For image classification, we adopt 5 popular UDA benchmarks of varied scales.
\emph{Office-31}~\cite{saenko2010adapting} is a small-scale benchmark with 31 classes in 3 domains: Amazon (A), DSLR (D), and Webcam (W).
\emph{Office-Home}~\cite{venkateswara2017Deep} is a medium-scale benchmark with 65 classes in 4 domains: Art (Ar), Clipart (Cl), Product (Pr), and Real-World (Re).
\emph{VisDA}~\cite{peng2017visda} is a large-scale benchmark with over 200k images across 12 classes in 2 domains: Training (T) and Validation (V). 
\emph{DomainNet}~\cite{peng2019moment} is a large-scale benchmark with 600k images. We take a subset of 126 classes with 7 tasks\cite{saito2019semi} from 4 domains: Real (R), Clipart (C), Painting (P), and Sketch (S).
\emph{Image-Sketch}~\cite{wang2019learning} is a large-scale benchmark with 1000 classes in 2 domains: ImageNet (I) and Sketch (S).
For semantic segmentation, we use \emph{Cityscapes}\cite{cordts2016cityscapes} as the target domain and either \emph{GTA5}\cite{richter2016playing} or \emph{SYNTHIA}~\cite{ros2016synthia} as the source.

\textbf{UDA methods.}
We evaluate calibration on 10 UDA methods across 5 UDA scenarios.
For image classification, we cover closed-set UDA methods (ATDOC~\cite{liang2021domain}, BNM~\cite{cui2020towards}, MCC~\cite{jin2020minimum}, CDAN~\cite{long2018conditional}, SAFN~\cite{xu2019larger}, MCD~\cite{saito2018maximum}), partial-set UDA methods (ATDOC~\cite{liang2021domain}, MCC~\cite{jin2020minimum}, PADA~\cite{cao2018partial}), the whit-box source-free UDA method (SHOT~\cite{liang2020we}), and the black-box source-free UDA method (DINE~\cite{liang2022dine}).
For semantic segmentation, we focus on calibrating source models without any adaptation.

\textbf{Calibration baselines.}
To provide a comprehensive comparison, we consider typical calibration baselines in UDA, including the no calibration baseline (No Calib.), IID calibration methods (MatrixScal~\cite{guo2017calibration}, VectorScal~\cite{guo2017calibration}, TempScal~\cite{guo2017calibration}), cross-domain calibration methods (CPCS~\cite{park2020calibrated}, TransCal~\cite{wang2020transferable}), and a general calibration method (Ensemble~\cite{lakshminarayanan2017simple}).

\textbf{Implementation details.}
We train all UDA models using the official code until convergence on a single RTX TITAN 16GB GPU. We adopt ResNet-101~\cite{resnet16} for \emph{VisDA} and segmentation tasks, ResNet-34~\cite{resnet16} for \emph{DomainNet}, and ResNet-50~\cite{resnet16} for all other tasks.
For PseudoCal, a fixed mix ratio $\lambda$ of $0.65$ is employed in all experiments. The UDA model is utilized for one-epoch inference with \textit{mixup} to generate the pseudo-target set. The reported results are averaged over five random runs.

\begin{table*}[!htbp]\centering
\caption{ECE ($\%$) of closed-set UDA on \emph{Office-Home} (\emph{Home}). Lower is better. \textbf{bold}: Best case.} 
\label{tab:uda-home}
\resizebox{1.0\textwidth}{!}{$
\begin{tabular}{l|rrrrd|rrrrd|rrrrda}
\cmidrule{0-15}
\multirow{2}{*}{Method} & \multicolumn{5}{c|}{ATDOC~\cite{liang2021domain}} & \multicolumn{5}{c|}{BNM~\cite{cui2020towards}} & \multicolumn{5}{c}{MCC~\cite{jin2020minimum}} \\
 & $\to$Ar &$\to$Cl &$\to$Pr &$\to$Re &avg &$\to$Ar &$\to$Cl &$\to$Pr &$\to$Re &avg &$\to$Ar &$\to$Cl &$\to$Pr &$\to$Re &avg  \\
\cmidrule{0-15}
No Calib. &10.07 &22.35 &8.61 &6.06 &11.77 &30.97 &39.85 &19.70 &16.73 &26.81 &13.25 &23.11 &12.33 &10.53 &14.81 \\
MatrixScal~\cite{guo2017calibration} &23.43 &33.59 &19.45 &18.45 &23.73 &28.58 &39.38 &21.27 &19.08 &27.07 &26.23 &36.10 &21.99 &20.50 &26.20 \\
VectorScal~\cite{guo2017calibration} &11.52 &23.33 &7.62 &6.56 &12.26 &22.86 &32.39 &14.70 &11.26 &20.30 &11.80 &23.58 &10.04 &8.57 &13.50 \\
TempScal~\cite{guo2017calibration} &6.19 &17.54 &\textbf{3.98} &\textbf{3.03} &7.68 &23.11 &30.32 &13.70 &10.25 &19.35 &6.74 &16.25 &\textbf{5.08} &4.10 &8.04 \\
CPCS~\cite{park2020calibrated} &14.13 &14.75 &11.02 &7.33 &11.81 &24.76 &25.02 &14.90 &8.80 &18.37 &19.11 &28.59 &14.65 &5.55 &16.97 \\
TransCal~\cite{wang2020transferable} &18.09 &6.52 &16.03 &18.29 &14.73 &17.44 &27.22 &9.14 &5.47 &14.82 &11.73 &3.86 &6.70 &8.16 &7.61 \\
Ensemble~\cite{lakshminarayanan2017simple} &7.38 &18.01 &5.51 &4.22 &8.78 &22.50 &30.68 &14.38 &12.53 &20.02 &9.76 &19.20 &9.48 &7.90 &11.58 \\
PseudoCal  &\textbf{2.42} &\textbf{2.93} &5.84 &5.07 &\textbf{4.07} &\textbf{17.34} &\textbf{16.03} &\textbf{6.20} &\textbf{4.68} &\textbf{11.06} &\textbf{2.85} &\textbf{2.25} &5.18 &\textbf{3.57} &\textbf{3.47} \\
\cmidrule{0-15}
Oracle &1.71 &1.91 &2.29 &1.69 &1.90 &2.20 &2.53 &2.36 &1.60 &2.17 &2.25 &1.64 &2.22 &1.91 &2.00 \\
Accuracy &66.42 &52.39 &76.60 &77.74 &68.29 &65.42 &53.69 &76.51 &78.98 &68.65 &61.03 &47.47 &72.37 &74.03 &63.73 \\
\cmidrule{0-15}
\toprule[1pt]
\multirow{2}{*}{Method} & \multicolumn{5}{c|}{CDAN~\cite{long2018conditional}} & \multicolumn{5}{c|}{SAFN~\cite{xu2019larger}} & \multicolumn{5}{c}{MCD~\cite{saito2018maximum}} & \emph{Home} \\
&$\to$Ar &$\to$Cl &$\to$Pr &$\to$Re &avg &$\to$Ar &$\to$Cl &$\to$Pr &$\to$Re &avg &$\to$Ar &$\to$Cl &$\to$Pr &$\to$Re &avg & AVG\\
\cmidrule{0-16}
No Calib. &13.38 &22.94 &12.15 &10.00 &14.62 &16.57 &27.90 &13.16 &11.93 &17.39 &16.36 &25.96 &13.29 &11.97 &16.89 &17.05 \\
MatrixScal~\cite{guo2017calibration} &24.22 &33.02 &20.85 &19.04 &24.28 &24.88 &36.22 &21.45 &19.76 &25.58 &28.53 &39.39 &24.35 &22.43 &28.67 &25.92 \\
VectorScal~\cite{guo2017calibration} &10.58 &20.67 &8.68 &7.72 &11.91 &12.57 &22.72 &9.80 &8.59 &13.42 &13.25 &23.35 &8.51 &8.37 &13.37 &14.13 \\
TempScal~\cite{guo2017calibration} &6.89 &15.44 &5.01 &4.19 &7.88 &6.99 &16.13 &4.56 &\textbf{4.07} &7.94 &6.01 &12.15 &\textbf{3.56} &\textbf{3.54} &6.31 &9.53 \\
CPCS~\cite{park2020calibrated}  &18.38 &33.56 &15.29 &9.90 &19.28 &14.98 &30.54 &10.06 &12.11 &16.92 &25.13 &27.26 &10.17 &14.29 &19.21 &17.09 \\
TransCal~\cite{wang2020transferable} &14.76 &4.72 &12.07 &13.73 &11.32 &3.50 &6.87 &\textbf{3.77} &4.15 &4.57 &10.78 &\textbf{2.66} &10.31 &11.27 &8.76 &10.30 \\
Ensemble~\cite{lakshminarayanan2017simple} &10.07 &18.58 &9.15 &7.23 &11.26 &14.82 &24.90 &11.17 &9.86 &15.19 &12.36 &20.87 &8.93 &7.64 &12.45 &13.21 \\
PseudoCal &\textbf{5.10} &\textbf{3.72} &\textbf{4.71} &\textbf{2.40} &\textbf{3.98} &\textbf{3.05} &\textbf{3.34} &6.86 &4.37 &\textbf{4.41} &\textbf{4.07} &2.86 &6.26 &3.72 &\textbf{4.23} &\textbf{5.20} \\
\cmidrule{0-16}
Oracle &3.61 &2.84 &2.26 &1.94 &2.66 &1.96 &2.48 &2.52 &1.74 &2.17 &2.65 &2.27 &2.30 &2.22 &2.36 &2.21\\
Accuracy &62.26 &49.99 &71.19 &73.79 &64.31 &65.84 &51.90 &73.78 &75.09 &66.66 &59.04 &46.80 &68.75 &71.39 &61.49 &65.52\\
\bottomrule[1pt]
\end{tabular}
$}
\end{table*}

\vspace{-5pt}

\begin{table*}[!htbp]\centering
\caption{ECE ($\%$) of closed-set UDA on \emph{Office-31} (\emph{Office}) and \emph{VisDA}.} 
\label{tab:uda-office-visda}
\resizebox{1.0\textwidth}{!}{$
\begin{tabular}{l|rrrdd|rrrdd|rrrddaa}
\cmidrule{0-15}
\multirow{2}{*}{Method} & \multicolumn{5}{c|}{ATDOC~\cite{liang2021domain}} & \multicolumn{5}{c|}{BNM~\cite{cui2020towards}} & \multicolumn{5}{c}{MCC~\cite{jin2020minimum}} \\
 &$\to$A &$\to$D &$\to$W &avg & T$\to$V &$\to$A &$\to$D &$\to$W &avg & T$\to$V &$\to$A &$\to$D &$\to$W &avg & T$\to$V  \\
\cmidrule{0-15}
No Calib. &12.17 &4.59 &6.66 &7.81 &10.38 &23.41 &11.12 &8.27 &14.27 &17.10 &19.29 &6.18 &7.80 &11.09 &17.42 \\
MatrixScal~\cite{guo2017calibration} &14.70 &9.58 &13.21 &12.49 &16.40 &23.29 &11.22 &8.81 &14.44 &18.77 &20.41 &9.70 &10.21 &13.44 &18.84 \\
VectorScal~\cite{guo2017calibration} &16.59 &4.57 &6.43 &9.20 &14.71 &24.20 &8.15 &4.11 &12.15 &16.64 &22.27 &5.12 &3.16 &10.18 &16.77 \\
TempScal~\cite{guo2017calibration} &22.39 &\textbf{3.39} &4.18 &9.99 &10.53 &23.85 &9.23 &4.98 &12.69 &13.72 &21.38 &3.79 &3.00 &9.39 &13.28 \\
CPCS~\cite{park2020calibrated}  &24.64 &7.98 &8.94 &13.85 &16.65 &22.45 &11.65 &\textbf{2.02} &12.04 &15.36 &30.16 &4.69 &3.03 &12.63 &7.14 \\
TransCal~\cite{wang2020transferable} &12.14 &14.21 &14.64 &13.67 &6.36 &14.86 &\textbf{5.22} &2.70 &7.59 &8.79 &6.53 &3.77 &3.91 &4.74 &12.21 \\
Ensemble~\cite{lakshminarayanan2017simple} &9.79 &3.60 &\textbf{4.09} &5.83 &8.53 &19.77 &6.92 &4.63 &10.44 &14.84 &17.48 &3.07 &4.88 &8.48 &15.32 \\
PseudoCal &\textbf{3.85} &6.64 &4.98 &\textbf{5.16} &\textbf{5.27} &\textbf{9.48} &6.30 &3.97 &\textbf{6.58} &\textbf{3.03} &\textbf{4.61} &\textbf{2.68} &\textbf{2.82} &\textbf{3.37} &\textbf{1.20} \\
\cmidrule{0-15}
Oracle &2.13 &2.49 &3.15 &2.59 &0.52 &2.52 &2.65 &1.40 &2.19 &0.93 &2.24 &2.36 &2.67 &2.42 &1.12 \\
Accuracy &73.23 &91.57 &88.93 &84.58 &75.96 &72.56 &88.35 &90.94 &83.95 &76.23 &69.69 &91.37 &89.06 &83.37 &78.00 \\
\cmidrule{0-15}
\toprule[1pt]
\multirow{2}{*}{Method} & \multicolumn{5}{c|}{CDAN~\cite{long2018conditional}} & \multicolumn{5}{c|}{SAFN~\cite{xu2019larger}} & \multicolumn{5}{c}{MCD~\cite{saito2018maximum}} & \emph{Office} & \emph{VisDA}  \\
&$\to$A &$\to$D &$\to$W &avg & T$\to$V &$\to$A &$\to$D &$\to$W &avg & T$\to$V &$\to$A &$\to$D &$\to$W &avg & T$\to$V &AVG &AVG \\
\cmidrule{0-17}
No Calib. &17.02 &9.34 &7.96 &11.44 &15.90 &21.34 &6.17 &6.68 &11.40 &18.53 &16.71 &9.49 &8.88 &11.69 &17.58 &11.28 &16.15 \\
MatrixScal~\cite{guo2017calibration} &19.16 &11.90 &14.91 &15.32 &18.70 &21.99 &9.49 &13.97 &15.15 &20.98 &19.07 &9.83 &13.49 &14.13 &19.88 &14.16 &18.93 \\
VectorScal~\cite{guo2017calibration} &21.52 &6.04 &3.60 &10.39 &17.13 &22.33 &3.22 &\textbf{2.20} &9.25 &19.21 &19.13 &5.87 &4.61 &9.87 &19.05 &10.17 &17.25 \\
TempScal~\cite{guo2017calibration} &18.54 &5.70 &3.41 &9.21 &14.19 &23.95 &3.21 &2.83 &9.99 &14.40 &25.37 &\textbf{3.44} &\textbf{2.36} &10.39 &10.22 &10.28 &12.72  \\
CPCS~\cite{park2020calibrated}  &17.47 &30.95 &5.67 &18.03 &15.45 &23.15 &8.21 &18.21 &16.52 &17.88 &27.69 &11.85 &19.01 &19.52 &10.56 &15.43 &13.84 \\
TransCal~\cite{wang2020transferable} &\textbf{4.84} &7.44 &6.84 &6.38 &4.07 &8.14 &\textbf{3.04} &2.81 &\textbf{4.67} &8.23 &5.13 &5.65 &4.76 &5.18 &\textbf{3.74} &7.04 &7.23 \\
Ensemble~\cite{lakshminarayanan2017simple} &10.92 &4.98 &3.29 &6.40 &13.30 &18.89 &3.81 &5.75 &9.48 &17.31 &14.56 &6.25 &5.49 &8.77 &14.82 &8.23 &14.02 \\
PseudoCal &6.58 &\textbf{4.78} &\textbf{3.04} &\textbf{4.80} &\textbf{3.04} &\textbf{4.13} &7.92 &5.51 &5.85 &\textbf{7.54} &\textbf{4.22} &5.97 &5.33 &\textbf{5.17} &6.71 &\textbf{5.16} &\textbf{4.46} \\
\cmidrule{0-17}
Oracle &3.21 &3.26 &2.17 &2.88 &1.00 &2.21 &2.90 &1.75 &2.29 &1.82 &2.11 &3.55 &1.76 &2.47 &0.99 &2.47 &1.06 \\
Accuracy &66.03 &87.15 &87.17 &80.12 &75.24 &68.95 &89.96 &88.55 &82.49 &73.91 &67.07 &86.14 &85.53 &79.58 &72.18 &82.35 &75.25 \\
\bottomrule[1pt]
\end{tabular}
$}
\end{table*}

\begin{table*}[!htbp]\centering
\caption{ECE ($\%$) of closed-set UDA on \emph{DomainNet} (\emph{DNet}).} 
\label{tab:uda-dmnt}
\resizebox{1.0\textwidth}{!}{$
\begin{tabular}{l|rrrrd|rrrrd|rrrrda}
\cmidrule{0-15}
\multirow{2}{*}{Method} & \multicolumn{5}{c|}{ATDOC~\cite{liang2021domain}} & \multicolumn{5}{c|}{BNM~\cite{cui2020towards}} & \multicolumn{5}{c}{MCC~\cite{jin2020minimum}} \\
 & $\to$C &$\to$P &$\to$R &$\to$S &avg &$\to$C &$\to$P &$\to$R &$\to$S &avg &$\to$C &$\to$P &$\to$R &$\to$S &avg  \\
\cmidrule{0-15}
No Calib. &9.54 &7.38 &3.75 &12.29 &8.24 &28.57 &22.10 &15.37 &31.27 &24.33 &8.63 &7.77 &4.79 &13.61 &8.70 \\
MatrixScal~\cite{guo2017calibration} &25.41 &22.91 &15.58 &31.17 &23.77 &29.00 &25.48 &18.31 &35.11 &26.98 &26.26 &23.84 &15.99 &33.25 &24.83 \\
VectorScal~\cite{guo2017calibration} &13.28 &9.86 &3.97 &17.73 &11.21 &20.91 &14.50 &9.88 &24.89 &17.55 &13.80 &10.71 &4.49 &19.69 &12.17 \\
TempScal~\cite{guo2017calibration} &8.69 &7.71 &1.94 &11.82 &7.54 &19.04 &13.62 &9.40 &20.30 &15.59 &8.38 &8.32 &\textbf{2.36} &13.88 &8.23 \\
CPCS~\cite{park2020calibrated} &10.78 &4.72 &4.46 &13.38 &8.34 &8.23 &7.92 &7.98 &9.29 &8.36 &9.03 &4.33 &3.44 &17.21 &8.50 \\
TransCal~\cite{wang2020transferable} &23.02 &24.76 &26.65 &19.68 &23.52 &\textbf{6.52} &\textbf{1.84} &\textbf{5.82} &9.39 &\textbf{5.89} &22.27 &24.06 &23.45 &18.03 &21.95 \\
Ensemble~\cite{lakshminarayanan2017simple} &6.32 &4.54 &\textbf{1.59} &9.05 &5.37 &23.44 &18.61 &12.61 &26.21 &20.22 &5.71 &5.10 &2.57 &10.34 &5.93 \\
PseudoCal  &\textbf{1.82} &\textbf{1.41} &2.51 &\textbf{1.70} &\textbf{1.86} &10.27 &6.01 &6.18 &\textbf{5.86} &7.08 &\textbf{1.35} &\textbf{1.89} &2.38 &\textbf{3.10} &\textbf{2.18}  \\
\cmidrule{0-15}
Oracle &1.55 &0.94 &0.86 &1.07 &1.10 &2.40 &1.66 &3.40 &1.30 &2.19 &1.16 &1.44 &1.09 &0.89 &1.14 \\
Accuracy &56.05 &60.64 &74.95 &52.08 &60.93 &56.62 &63.13 &74.30 &52.25 &61.57 &50.89 &57.74 &71.62 &46.39 &56.66 \\
\cmidrule{0-15}
\toprule[1pt]
\multirow{2}{*}{Method} & \multicolumn{5}{c|}{CDAN~\cite{long2018conditional}} & \multicolumn{5}{c|}{SAFN~\cite{xu2019larger}} & \multicolumn{5}{c}{MCD~\cite{saito2018maximum}} & \emph{DNet} \\
&$\to$C &$\to$P &$\to$R &$\to$S &avg &$\to$C &$\to$P &$\to$R &$\to$S &avg &$\to$C &$\to$P &$\to$R &$\to$S &avg & AVG\\
\cmidrule{0-16}
No Calib. &10.17 &9.64 &5.56 &14.44 &9.95 &17.94 &14.44 &10.15 &21.26 &15.95 &9.56 &7.40 &3.80 &12.93 &8.42 &12.60 \\
MatrixScal~\cite{guo2017calibration} &24.96 &23.05 &15.80 &31.26 &23.77 &21.80 &18.69 &11.16 &29.89 &20.38 &19.94 &16.38 &10.05 &28.23 &18.65 &23.06 \\
VectorScal~\cite{guo2017calibration} &11.88 &9.63 &4.37 &17.67 &10.89 &15.78 &10.45 &4.73 &20.32 &12.82 &14.56 &10.24 &5.75 &20.29 &12.71 &12.89 \\
TempScal~\cite{guo2017calibration} &7.92 &8.31 &2.75 &12.30 &7.82 &9.61 &8.15 &4.12 &14.18 &9.02 &6.48 &6.96 &4.06 &11.20 &7.18 &9.23 \\
CPCS~\cite{park2020calibrated}  &10.75 &4.28 &5.57 &6.91 &6.88 &10.92 &5.91 &8.22 &22.59 &11.91 &7.02 &3.51 &1.96 &21.79 &8.57 &8.76 \\
TransCal~\cite{wang2020transferable} &20.92 &21.41 &22.93 &16.93 &20.55 &10.75 &12.88 &14.28 &6.88 &11.20 &21.48 &24.99 &27.45 &18.95 &23.22 &17.72 \\
Ensemble~\cite{lakshminarayanan2017simple} &7.21 &6.74 &3.54 &11.29 &7.20 &16.59 &13.25 &9.08 &19.52 &14.61 &7.25 &5.27 &2.86 &11.34 &6.68 &10.00 \\
PseudoCal &\textbf{1.58} &\textbf{1.89} &\textbf{1.86} &\textbf{2.67} &\textbf{2.00} &\textbf{3.33} &\textbf{1.30} &\textbf{1.50} &\textbf{2.76} &\textbf{2.22} &\textbf{2.27} &\textbf{1.16} &\textbf{1.01} &\textbf{1.70} &\textbf{1.53} &\textbf{2.81} \\
\cmidrule{0-16}
Oracle &1.45 &1.08 &1.07 &0.94 &1.13 &1.43 &0.92 &1.21 &0.72 &1.07 &1.33 &0.97 &0.56 &0.68 &0.88 &1.25 \\
Accuracy &53.11 &59.13 &71.82 &49.09 &58.29 &49.59 &58.03 &66.40 &47.66 &55.42 &48.85 &57.99 &65.32 &47.95 &55.03 &57.98\\
\bottomrule[1pt]
\end{tabular}
$}
\end{table*}

\subsection{Results}
We evaluate the calibration performance of PseudoCal across 5 UDA scenarios. For classification tasks, we report the average ECE results for UDA tasks with the same target domain in Tables~\ref{tab:uda-home}-\ref{tab:sfda-dmnt}. For segmentation tasks, we take each pixel as a sample and report the results in Table~\ref{tab:seg}. `Oracle' refers to the aforementioned `Oracle' calibration with target labels, and `Accuracy' ($\%$) denotes the target accuracy of the UDA model.

\textbf{Closed-set UDA.} We evaluate 6 UDA methods on 4 benchmarks for closed-set UDA. Specifically, we report the ECE for \emph{Office-Home} in Table~\ref{tab:uda-home}, ECE for both \emph{Office-31} and \emph{VisDA} in Table~\ref{tab:uda-office-visda}, and ECE for \emph{DomainNet} in Table~\ref{tab:uda-dmnt}. 
PseudoCal consistently achieves a low ECE close to `Oracle', significantly outperforming other calibration methods by a wide margin. On the evaluated benchmarks, PseudoCal shows average ECE improvements of $4.33\%$ on \emph{Office-Home}, $1.88\%$ on \emph{Office-31}, $2.77\%$ on \emph{VisDA}, and $5.95\%$ on \emph{DomainNet} when compared to the second-best calibration method.

\textbf{Partial-set UDA.} We evaluate 3 partial-set UDA methods on \emph{Office-Home} and report the ECE in Table~\ref{tab:pda-home}. PseudoCal consistently performs the best on average and outperforms the second-best method (Ensemble) by a significant margin of $4.24\%$.

\textbf{Source-free UDA.} We evaluate the popular source-free UDA settings using SHOT for the white-box setting and DINE for the black-box setting. We report the ECE for large-scale benchmarks \emph{DomainNet} and \emph{Image-Sketch} together in Table~\ref{tab:sfda-dmnt} and compare PseudoCal with the other source-free method Ensemble. PseudoCal outperforms Ensemble on both benchmarks by significant margins, with $7.44\%$ on \emph{DomainNet} and $15.05\%$ on \emph{Image-Sketch}.

\textbf{Semantic segmentation.} In addition to classification tasks, we evaluate PseudoCal on domain adaptive semantic segmentation tasks and report the ECE in Table~\ref{tab:seg}. PseudoCal performs the best on average and demonstrates an average ECE improvement of $4.62\%$ over the no-calibration baseline.

\begin{table*}[!htbp]\centering
\caption{ECE ($\%$) of partial-set UDA on \emph{Office-Home} (\emph{Home}).} 
\label{tab:pda-home}
\resizebox{1.0\textwidth}{!}{$
\begin{tabular}{l|rrrrd|rrrrd|rrrrd|a}
\toprule[1pt]
\multirow{2}{*}{Method} & \multicolumn{5}{c|}{ATDOC~\cite{liang2021domain}} & \multicolumn{5}{c|}{MCC~\cite{jin2020minimum}} & \multicolumn{5}{c|}{PADA~\cite{cao2018partial}}& \emph{Home} \\
 & $\to$Ar &$\to$Cl &$\to$Pr &$\to$Re &avg &$\to$Ar &$\to$Cl &$\to$Pr &$\to$Re &avg &$\to$Ar &$\to$Cl &$\to$Pr &$\to$Re &avg &AVG  \\
\cmidrule{0-16}
No Calib. &16.68 &28.47 &20.00 &12.26 &19.35 &12.71 &22.17 &12.21 &8.99 &14.02 &9.45 &19.09 &9.19 &6.77 &11.13 &14.83 \\
MatrixScal~\cite{guo2017calibration} &23.67 &35.94 &22.17 &16.73 &24.63 &24.40 &34.69 &21.71 &18.32 &24.78 &27.56 &36.51 &24.57 &20.55 &27.30 &25.57 \\
VectorScal~\cite{guo2017calibration} &16.50 &28.49 &16.13 &10.48 &17.90 &12.74 &22.08 &10.64 &8.72 &13.54 &14.59 &24.11 &9.90 &9.17 &14.44 &15.29 \\
TempScal~\cite{guo2017calibration} &13.40 &24.79 &14.91 &8.72 &15.45 &7.12 &15.97 &6.04 &\textbf{4.35} &8.37 &8.92 &18.20 &6.21 &4.08 &9.35 &11.06 \\
CPCS~\cite{park2020calibrated} &19.39 &29.74 &13.86 &14.63 &19.41 &12.73 &28.11 &9.09 &10.69 &15.16 &24.40 &22.74 &17.30 &27.67 &23.03 &19.20 \\
TransCal~\cite{wang2020transferable} &10.64 &\textbf{5.17} &\textbf{5.88} &11.30 &8.25 &9.44 &4.27 &\textbf{5.41} &6.98 &6.53 &22.70 &11.00 &23.00 &26.77 &20.87 &11.88 \\
Ensemble~\cite{lakshminarayanan2017simple} &11.98 &21.28 &13.44 &8.62 &13.83 &9.22 &18.54 &10.11 &6.78 &11.16 &5.30 &11.86 &\textbf{4.43} &\textbf{3.92} &\textbf{6.38} &10.46 \\
PseudoCal  &\textbf{7.87} &10.90 &6.24 &\textbf{4.83} &\textbf{7.46} &\textbf{3.74} &\textbf{3.63} &6.93 &4.81 &\textbf{4.78} &\textbf{4.72} &\textbf{3.45} &10.77 &6.69 &6.41 &\textbf{6.22} \\
\cmidrule{0-16}
Oracle &4.13 &4.45 &4.37 &4.08 &4.26 &2.81 &3.01 &3.06 &2.37 &2.81 &3.94 &2.65 &4.80 &3.03 &3.61 &3.56 \\
Accuracy &63.02 &50.70 &65.92 &73.71 &63.34 &65.53 &51.68 &73.41 &78.23 &67.21 &55.65 &44.06 &61.23 &66.54 &56.87 &62.47 \\
\bottomrule[1pt]
\end{tabular}
$}
\end{table*}

\begin{table*}[!htbp]\centering
\caption{ECE ($\%$) of source-free UDA on \emph{DomainNet} (\emph{DNet}) and \emph{ImageNet-Sketch} (\emph{Sketch}).} 
\label{tab:sfda-dmnt}
\resizebox{1.0\textwidth}{!}{$
\begin{tabular}{l|rrrrdd|rrrrdd|a|a}
\toprule[1pt]
\multirow{2}{*}{Method} & \multicolumn{6}{c|}{SHOT~\cite{liang2020we}} & \multicolumn{6}{c|}{DINE~\cite{liang2022dine}} & \emph{DNet} & \emph{Sketch} \\
 &$\to$C &$\to$P &$\to$R &$\to$S &avg & I$\to$S&$\to$C &$\to$P &$\to$R &$\to$S &avg & I$\to$S &AVG &AVG  \\
\cmidrule{0-14}
No Calib. &17.16 &21.19 &10.03 &23.14 &17.88 &34.71 &21.99 &22.51 &12.39 &30.34 &21.81 &58.85 &19.84 &46.78 \\
Ensemble~\cite{lakshminarayanan2017simple} &14.24 &17.94 &7.81 &19.49 &14.87 &33.03 &17.88 &18.86 &10.83 &25.33 &18.22 &53.24 &16.54 &43.14 \\
PseudoCal  &\textbf{6.66} &\textbf{7.78} &\textbf{2.91} &\textbf{6.67} &\textbf{6.00} &\textbf{8.42} &\textbf{14.42} &\textbf{12.95} &\textbf{5.30} &\textbf{16.15} &\textbf{12.20} &\textbf{47.76} &\textbf{9.10} &\textbf{28.09} \\
\cmidrule{0-14}
Oracle &3.27 &2.52 &1.37 &2.18 &2.33 &4.39 &1.75 &1.80 &1.29 &1.37 &1.55 &5.90 &1.94 &5.14 \\
Accuracy &66.52 &64.48 &78.34 &59.64 &67.25 &34.29 &63.76 &65.47 &80.69 &55.51 &66.36 &22.27 &66.80 &28.28 \\
\bottomrule[1pt]
\end{tabular}
$}
\end{table*}

\begin{figure}[!htbp]
    \begin{minipage}[!htbp]{0.45\textwidth}
        \vspace{31.5pt}
        \centering
        \resizebox{1.0\textwidth}{!}{
        \begin{tabular}{l|r|r|a}
        \toprule[1pt]
        Method &\emph{GTA5~\cite{richter2016playing}} &\emph{SYNTHIA~\cite{ros2016synthia}} & AVG \\
        \cmidrule{0-3}
        No Calib. &7.87 &23.08 & 15.48 \\
        TempScal~\cite{guo2017calibration} &4.61 &19.24 &11.93 \\
        Ensemble~\cite{lakshminarayanan2017simple}  &\textbf{2.66} &20.84 & 11.75 \\
        PseudoCal &5.73 &\textbf{15.99} & \textbf{10.86} \\
        \cmidrule{0-3}
        Oracle &0.52 &4.5 & 2.51 \\
        \bottomrule[1pt]
        \end{tabular}
        }
        \vspace{3pt}
        \captionof{table}{ECE ($\%$) of Semantic segmentation.}
        \label{tab:seg}
    \end{minipage}
    \hfill
    \begin{minipage}[!htbp]{0.45\textwidth}
        \vspace{0pt}
        \centering
        \resizebox{1.0\textwidth}{!}{
        \begin{tabular}{l|r|r|r}
        \toprule[1pt]
        Method &ECE~\cite{guo2017calibration} ($\%$) & BS~\cite{brier1950verification} & NLL~\cite{goodfellow2016deep}\\
        \cmidrule{0-3}
        No Calib. & 11.52 & 0.5674 & 1.9592\\
        MatrixScal~\cite{guo2017calibration} &24.83 &0.6968 & 3.0858  \\
        VectorScal~\cite{guo2017calibration}  &10.77 & 0.5560 & 1.9154 \\
        TempScal~\cite{guo2017calibration} &10.63 & 0.5647 & 1.9418  \\
        CPCS~\cite{park2020calibrated} &5.48  & 0.5579 & 1.8781  \\
        TransCal~\cite{wang2020transferable} &23.38 & 0.6279 & 2.1089   \\
        Ensemble~\cite{lakshminarayanan2017simple} &10.08 & 0.5618 & 1.9260   \\
        PseudoCal &\textbf{3.63} & \textbf{0.5553} & \textbf{1.8697} \\
        \cmidrule{0-3}
        Oracle &1.29 & 0.5519 & 1.8597 \\
        \bottomrule[1pt]
        \end{tabular}
        }
        \vspace{3pt}
        \captionof{table}{ViT results of MCC~\cite{jin2020minimum} on C$\to$S.}
        \label{tab:vit}
    \end{minipage}
\end{figure}

\subsection{Discussions}
\begin{figure}[!htbp]
	\centering
	\footnotesize
	\setlength\tabcolsep{0.5 mm}
	\renewcommand\arraystretch{0.05}
	\begin{tabular}{cccc}
	   \includegraphics[width=0.25\linewidth,trim={0.0cm 0.0cm 0.0cm 0.0cm}, clip]{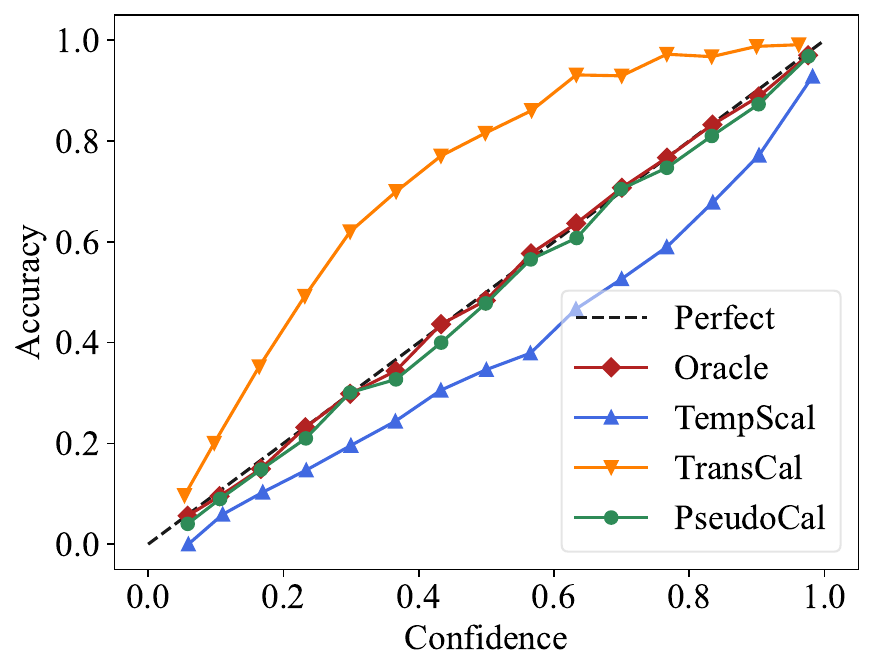} &
        \includegraphics[width=0.25\linewidth,trim={0.0cm 0.0cm 0.0cm 0.0cm}, clip]{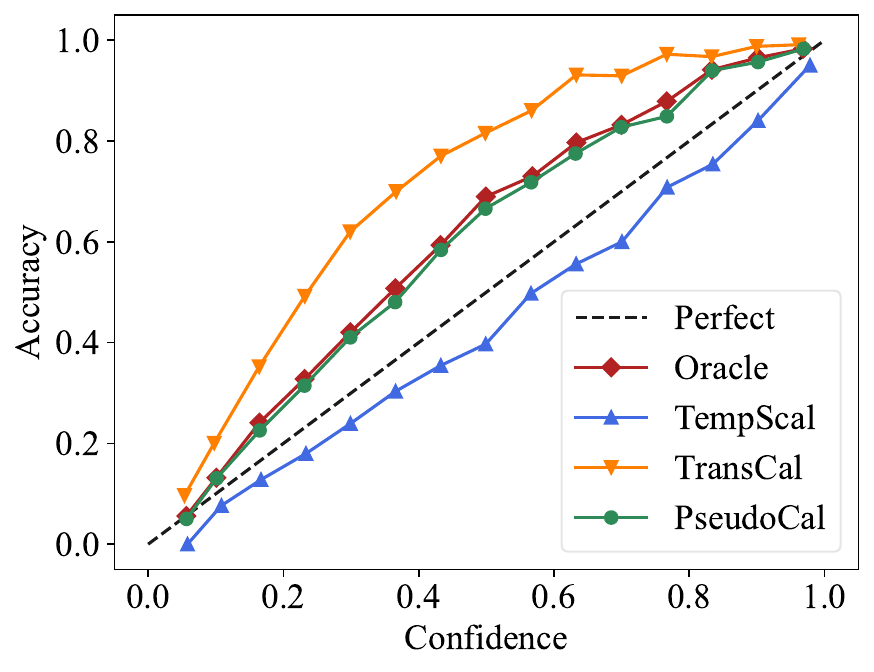} &
        \includegraphics[width=0.25\linewidth,trim={0.0cm 0.0cm 0.0cm 0.0cm}, clip]{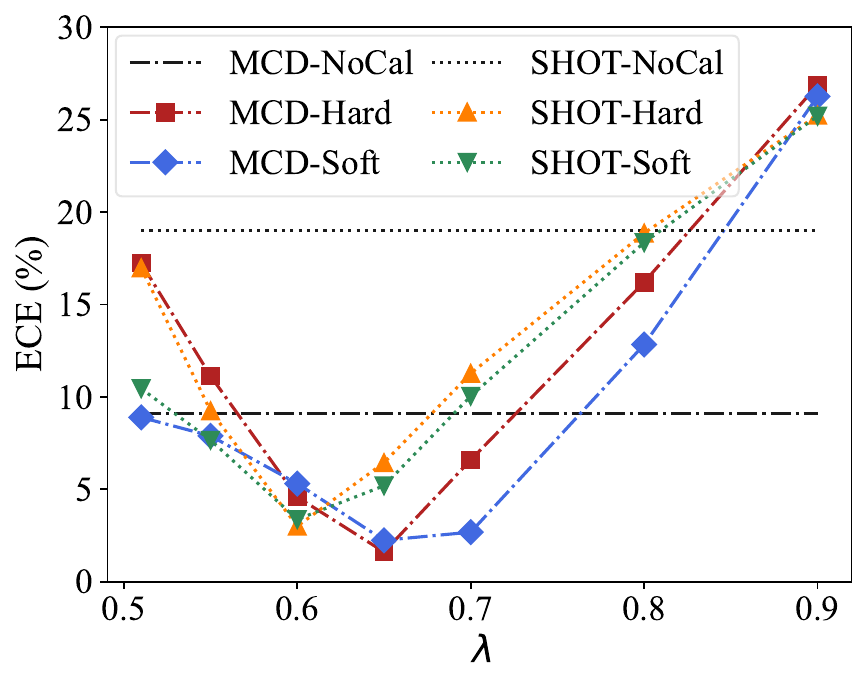} &
        \includegraphics[width=0.25\linewidth,trim={0.0cm 0.0cm 0.0cm 0.0cm}, clip]{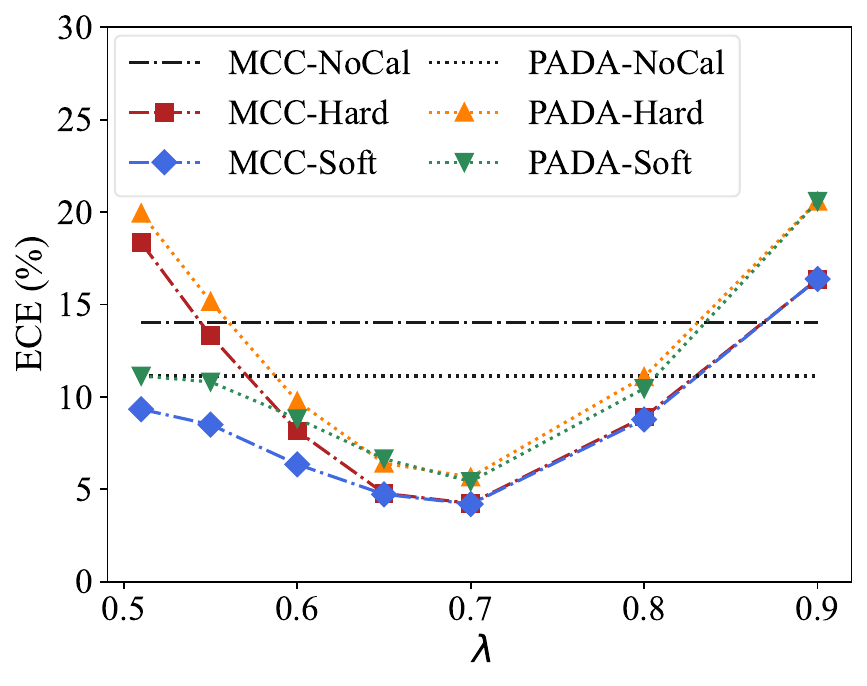} \\
        (a) CDAN \cite{long2018conditional} on R $\to$ S & (b) PADA \cite{cao2018partial} on Ar $\to$ Cl & (c) Closed-set on \textit{DNet} & (d) Partial-set on \textit{Home} \\
	\end{tabular}
	\caption{(a) and (b) provide the reliability diagrams of various calibration methods for a qualitative comparison. (c) and (d) present the sensitivity analysis of the fixed mix ratio $\lambda$.}
	\label{fig:ablation}
\end{figure} 

\textbf{Qualitative comparisons.} We present reliability diagrams~\cite{guo2017calibration} of different calibration methods in Figure~\ref{fig:ablation} (a)-(b). 
PseudoCal consistently aligns with `Oracle'in both UDA settings, while the state-of-the-art method TransCal deviates significantly.

\textbf{Impact of mix ratio $\lambda$.} We investigate the effect of the fixed mix ratio $\lambda$ used in \textit{mixup}, ranging from $0.51$ to $0.9$, on the ECE of two closed-set UDA methods (including SHOT) on \emph{DomainNet} in Figure~\ref{fig:ablation} (c) and two partial-set UDA methods on \emph{Office-Home} in Figure~\ref{fig:ablation} (d). We examine the $mixup$ with both `Hard' labels (one-hot label), and `Soft' labels (soft predictions). We find that PseudoCal achieves optimal performance within a medium range of $\lambda$ values, specifically between $0.6$ and $0.7$, regardless of the use of hard or soft labels. A $\lambda$ closer to $0.5$ generates more ambiguous samples, leading to increased wrong predictions, while a $\lambda$ closer to $1.0$ results in the opposite effect. To ensure simplicity, we adopt a value of $0.65$ for $\lambda$ with hard labels for all experiments.

\textbf{Robustness to backbones and metrics.} In order to examine the robustness of PseudoCal across different backbones and calibration metrics, we assess its performance using ViT-B~\cite{dosovitskiy2020image} as the backbone and present the results for three metrics in Table~\ref{tab:vit}. The findings reveal that PseudoCal consistently achieves top performance regardless of the choice of backbone or calibration metric.

\textbf{Impact of pseudo label quality.} Despite the low accuracy of pseudo labels (approximately $30\%$) on the `I$\to$S' task in Table~\ref{tab:sfda-dmnt}, PseudoCal consistently exhibits strong calibration performance, indicating its effectiveness even in the presence of low-quality pseudo labels.

\textbf{Ablation study on pseudo-target synthesis.} In our PseudoCal method, we utilize input-level \textit{mixup} with a fixed mix ratio ($\lambda$) to synthesize a pseudo-target sample by combining a pair of real samples with different pseudo labels. To conduct a thorough ablation study, we compare this data synthesis strategy with alternative choices, such as \textit{mixup} between samples with the same pseudo label (referred to as PseudoCal-same), instance-based augmentations~\cite{cubuk2020randaugment, chen2020improved}, mixing at different levels~\cite{yun2019cutmix, verma2019manifold}, using $\lambda$ values sampled from $\text{Beta}(0.3, 0.3)$~\cite{zhang2018mixup}, and directly utilizing pseudo-labeled real target samples~\cite{lee2013pseudo, sohn2020fixmatch}. A comprehensive comparison of all strategies is presented in Table~\ref{tab:abla-mix}. Our PseudoCal consistently outperforms the alternative options, benefiting from its superiority in accurately approximating the accuracy-confidence distribution of real target data.

\begin{table*}[!htbp]\centering
\caption{ECE ($\%$) of ablation expriments on pseudo-target synthesis.} 
\label{tab:abla-mix}
\resizebox{0.88\textwidth}{!}{$
\begin{tabular}{l|rr|rr|r|r|r|rr}
\toprule[1pt]
\multirow{2}{*}{Method} & 
\multicolumn{2}{c|}{MCD~\cite{saito2018maximum}} & 
\multicolumn{2}{c|}{BNM~\cite{cui2020towards}}&
CDAN~\cite{long2018conditional}&
SHOT~\cite{liang2020we}& 
DINE~\cite{liang2022dine}&
\multicolumn{2}{c}{PADA~\cite{cao2018partial}}
\\
 &D$\to$A &W$\to$A & Cl$\to$Pr &Pr$\to$Re & R$\to$C & I$\to$S& R$\to$C  &Ar$\to$Cl & Re$\to$Ar  \\
\cmidrule{0-9}
No Calib. &16.39 &17.03 &22.09 &15.72 &9.83 &34.71 &21.43 &20.35 &8.31 \\
MocoV2Aug~\cite{chen2020improved} &16.85 &17.21 &20.51 &14.98 &15.49 &28.63 &21.69 &25.81 &15.17 \\
RandAug~\cite{cubuk2020randaugment} &12.87 &11.53 &19.24 &11.37 &13.33 &29.28 &24.21 &18.47 &10.32 \\
CutMix~\cite{yun2019cutmix} &8.20 &6.39 &14.82 &10.60 &7.60 &23.18 &18.29 &15.96 &6.04 \\
ManifoldMix~\cite{verma2019manifold} &19.49 &19.27 &23.29 &16.94 &27.00 &50.54 &30.23 &36.04 &21.29 \\
Mixup-Beta~\cite{zhang2018mixup} &14.96 &13.11 &15.65 &11.24 &15.84 &26.74 &21.12 &23.85 &11.46 \\
Pseudo-Label~\cite{lee2013pseudo} &32.47 &33.35 &26.31 &19.65 &47.02 &65.7 &35.48 &56.18 &36.27 \\
Filtered-PL~\cite{sohn2020fixmatch} &31.74 &32.73 &26.14 &19.46 &45.35 &64.29 &34.93 &54.83 &35.1 \\
PseudoCal-same &19.31 &20.54 &22.50 &15.63 &25.43 &45.54 &27.58 &30.30 &18.46 \\
PseudoCal  &\textbf{4.38} &\textbf{4.06} &\textbf{6.31} &\textbf{4.76} &\textbf{1.51} &\textbf{8.42} &\textbf{13.71} &\textbf{2.95} &\textbf{3.71}\\
\cmidrule{0-9}
Oracle &2.31 &1.90 &3.14 &1.10 &1.28 &4.39 &1.62 &2.16 &2.87 \\
Accuracy &67.52 &66.63 &73.69 &80.35 &52.98 &34.29 &64.52 &43.82 &63.73 \\
\bottomrule[1pt]
\end{tabular}
$}
\end{table*}

\begin{table*}[!htbp]\centering
\caption{Correspondence between the real target and pseudo target. ECE ($\%$) calibration results.} 
\label{tab:correspond_discuss}
\resizebox{0.99\textwidth}{!}{$
\begin{tabular}{l|rr|rr|r|r|r|rr}
\toprule[1pt]
\multirow{2}{*}{Method} & 
\multicolumn{2}{c|}{MCD~\cite{saito2018maximum}} & 
\multicolumn{2}{c|}{BNM~\cite{cui2020towards}}&
CDAN~\cite{long2018conditional}&
SHOT~\cite{liang2020we}& 
DINE~\cite{liang2022dine}&
\multicolumn{2}{c}{PADA~\cite{cao2018partial}}
\\
 &D$\to$A &W$\to$A & Cl$\to$Pr &Pr$\to$Re & R$\to$C & I$\to$S& R$\to$C  &Ar$\to$Cl & Re$\to$Ar  \\
\cmidrule{0-9}
No Calib. &16.39 &17.03 &22.09 &15.72 &9.83 &34.71 &21.43 &20.35 &8.31 \\
PseudoCal  &\textbf{4.38} &\textbf{4.06} &\textbf{6.31} &\textbf{4.76} &\textbf{1.51} &\textbf{8.42} &\textbf{13.71} &\textbf{2.95} &\textbf{3.71}\\
\cmidrule{0-9}
Oracle &2.31 &1.90 &3.14 &1.10 &1.28 &4.39 &1.62 &2.16 &2.87 \\
Accuracy ($\%$) &67.52 &66.63 &73.69 &80.35 &52.98 &34.29 &64.52 &43.82 &63.73 \\
Correspond. ($\%$)  &61.94 &62.55 &61.53 &63.27 &61.91 &65.69 &63.94 &63.99 &60.68 \\
\bottomrule[1pt]
\end{tabular}
$}
\end{table*}

\textbf{Illustration of the real-pseudo correspondence.} In Figure~\ref{fig:atdoc-fig} (b), we present an illustration that highlights the remarkable similarity in the accuracy-confidence distribution between the real target and pseudo target. To provide a more comprehensive understanding of the correspondence, we delve into the sample-level analysis. Within each pair of real samples in the \textit{mixup} operation, we establish a correspondence when both the mixed pseudo sample and its dominant real sample are either correctly predicted or incorrectly predicted, evaluated by their respective labels. 
To quantify the observed correspondence, we calculate the correspondence rate as a percentage by dividing the number of corresponding pairs by the total number of pseudo-target samples. The results of our evaluation, presented in Table~\ref{tab:correspond_discuss}, demonstrate that PseudoCal consistently exhibits a high correspondence rate exceeding $60\%$ across different tasks with varied model accuracy. These findings provide further direct evidence in support of the existence of real-pseudo correspondence.

\textbf{Comparison with Ensemble.} We compare PseudoCal with a general calibration method Ensemble, which involves averaging predictions from multiple independently trained models. Our comparison demonstrates that Ensemble and PseudoCal are the only two methods that consistently maintain stable calibration performance across different UDA tasks. Notably, PseudoCal further surpasses Ensemble in terms of performance gains and computational efficiency.

\textbf{Limitations and broader impacts.}
PseudoCal has the following limitations and potential negative societal impacts:
(\textit{i}) Like other calibration methods compared, PseudoCal may occasionally increase ECE when the initial ECE is already small (see $\to$D in Table~\ref{tab:uda-office-visda}), which raises risks for safety-critical decision-making systems.
(\textit{ii}) While PseudoCal can handle the source-free calibration setting, it may face challenges in extreme cases with very few available target samples, such as only a single target sample.
(\textit{iii}) PseudoCal is partly dependent on the \textit{cluster assumption}, and it may fail if the target pseudo label is extremely poor, i.e., performing similarly to random trials.
(\textit{iv}) PseudoCal is based on \textit{temperature scaling} and may not be suitable for open-set settings where the confidence of unknown samples is determined by various thresholding methods rather than differential softmax.

\section{Conclusion}
\label{sec:conclusion}
In conclusion, we have introduced PseudoCal, a novel source-free calibration method for addressing the challenge of predictive uncertainty calibration in unsupervised domain adaptation (UDA). By relying solely on unlabeled target data, PseudoCal treats UDA calibration as an unsupervised calibration problem, distinguishing it from previous approaches based on the covariate shift assumption. Through the generation of a labeled pseudo-target set that replicates the accuracy-confidence distribution of real target samples, PseudoCal effectively converts the unsupervised calibration problem into a supervised one, leveraging popular IID methods such as \textit{temperature scaling} for calibration. Our comprehensive evaluations across diverse UDA settings, including source-free scenarios and semantic segmentation, consistently demonstrate the superior performance of PseudoCal compared to existing calibration methods. Notably, PseudoCal stands out in terms of both its simplicity and effectiveness, offering a promising solution for enhancing the calibration of UDA models in practical applications.

\newpage
{\small
\bibliographystyle{neurips}
\bibliography{reference}
}
\newpage
\appendix
\section{Algorithm}
The PyTorch-style pseudocode for our validation method PseudoCal is provided in Algorithm~\ref{pseudocode}.
\label{alg}
\begin{algorithm}[ht]
   \caption{PyTorch-style pseudocode for PseudoCal.}
   \label{pseudocode}
   
    \definecolor{codeblue}{rgb}{0.25,0.5,0.5}
    \lstset{
      basicstyle=\fontsize{9.6pt}{9.6pt}\ttfamily\bfseries,
      commentstyle=\fontsize{9.6pt}{9.6pt}\color{codeblue},
      keywordstyle=\fontsize{9.6pt}{9.6pt},
    }
\begin{lstlisting}[language=python]
# x: A batch of real target images with shuffled order.
# lam: The mix ratio, a fixed scalar value between 0.5 and 1.0.
# net: A trained UDA model in the evaluation mode.

# Perform pseudo-target synthesis for a mini-batch.
def pseudo_target_synthesis(x, lam, net):

    # Random batch index.
    rand_idx = torch.randperm(x.shape[0])
    inputs_a = x
    inputs_b = x[rand_idx]
    
    # Obtain model predictions and pseudo labels (pl).
    pred_a = net(inputs_a)
    pl_a = pred_a.max(dim=1)[1] 
    pl_b = pl_a[rand_idx]

    # Select the samples with distinct labels for the mixup.
    diff_idx = (pl_a != pl_b).nonzero(as_tuple=True)[0]
    
    # Mixup with images and labels.
    pseudo_inputs = lam * inputs_a + (1 - lam) * inputs_b
    if lam > 0.5:
        pseudo_labels = pl_a
    else:
        pseudo_labels = pl_b
        
    return pseudo_inputs[diff_idx], pseudo_labels[diff_idx]

# Perform supervised calibration using pseudo-target data.
def pseudoCal(pseudo_inputs, pseudo_labels, net):

    # Obtain predictions for the pseudo-target samples.
    pseudo_pred = net(pseudo_inputs)

    # Apply temperature scaling to estimate the
    # pseudo-target temperature as the real temperature.
    calib_method = TempScaling()
    pseudo_temp = calib_method(pseudo_pred, pseudo_labels)
    
    return pseudo_temp
\end{lstlisting}
\end{algorithm}

\section{Semantic Segmentation Calibration Details}
For our calibration experiments on semantic segmentation, we calibrate the models trained solely on the source domain (GTA5~\cite{richter2016playing} or SYNTHIA~\cite{ros2016synthia}) without any target adaptation. We treat each pixel as an individual sample in classification tasks for both \textit{mixup} and \textit{temperature scaling}. To address the computational complexity, we adopt the evaluation strategy suggested in previous studies~\cite{de2023reliability} and randomly sample 20,000 pixels from each image (with resolutions such as 1920*720) for calibration.

\section{Additional Calibration Metrics}
In addition to the Expected Calibration Error (ECE)~\cite{guo2017calibration} discussed in the main text, we also consider two other calibration metrics as follows. Let $\mathbf{y}_i$ represent the one-hot ground truth encoding for input sample $x_i$, and $\mathbf{\hat{p}}_i$ denote the predicted probability vector output by the model $\phi$.

\textbf{Negative Log-Likelihood (NLL)}~\cite{goodfellow2016deep} is also known as the cross-entropy loss. The NLL loss for a single sample $x_i$ is given by:
\begin{equation*}
\label{eq: nll}
\begin{aligned}
\mathcal{L}_{\rm NLL} = - \sum_{c=1}^C \mathbf{y}_i^c \log  \mathbf{\hat{p}}_i^c\\
 \end{aligned}
\end{equation*}

\textbf{Brier Score (BS)}~\cite{brier1950verification} can be defined as the squared error between the predicted probability vector and the one-hot label vector. The Brier Score for a single sample $x_i$ is given by:
\begin{equation*}
\label{eq: bs}
\begin{aligned}
\mathcal{L}_{\rm BS} =  \frac{1}{C} \sum_{c=1}^C (\mathbf{\hat{p}}_i^c - \mathbf{y}_i^c)^2\\
 \end{aligned}
\end{equation*}

In addition to the ViT results presented in the main text, we have observed consistent advantages of our PseudoCal method over existing calibration methods across all three calibration metrics: ECE, NLL, and BS. We choose to report the ECE results for most of the experiments as ECE~\cite{guo2017calibration} is one of the widely used calibration metrics.

\section{Full Calibration Results}
Due to space constraints in the main text, we have presented the average ECE results for tasks with the same target domain. For detailed calibration results of each task, please refer to Table~\ref{tab:full-uda-atdoc-home} to Table~\ref{tab:full-sfda-dine-dmnt}.

\begin{table*}[!htbp]\centering
\caption{ECE ($\%$) of a closed-set UDA method ATDOC~\cite{liang2021domain} on \emph{Office-Home}.} 
\label{tab:full-uda-atdoc-home}
\resizebox{1.0\textwidth}{!}{$
\begin{tabular}{l|rrrrrrrrrrrr|a}
\toprule[1pt]
Method &Ar $\to$ Cl &Ar $\to$ Pr &Ar $\to$ Re &Cl $\to$ Ar &Cl $\to$ Pr &Cl $\to$ Re &Pr $\to$ Ar &Pr $\to$ Cl &Pr $\to$ Re &Re $\to$ Ar &Re $\to$ Cl &Re $\to$ Pr & AVG \\
\cmidrule{0-13}
No Calib. &22.83 &10.57 &6.31 &10.77 &8.88 &6.38 &10.39 &22.61 &5.49 &9.06 &21.61 &6.38 &11.77 \\
MatrixScal~\cite{guo2017calibration} &35.03 &20.72 &18.28 &27.54 &24.73 &23.40 &22.51 &32.85 &13.66 &20.25 &32.89 &12.90 &23.73 \\
VectorScal~\cite{guo2017calibration} &22.05 &10.09 &5.85 &11.51 &7.74 &6.01 &15.12 &26.85 &7.81 &7.94 &21.10 &5.03 &12.26 \\
TempScal~\cite{guo2017calibration} &14.69 &5.55 &\textbf{2.60} &4.27 &\textbf{3.17} &\textbf{1.45} &9.67 &22.55 &5.04 &4.63 &15.37 &\textbf{3.21} &7.68 \\
CPCS~\cite{park2020calibrated} &8.37 &9.32 &6.44 &12.94 &14.94 &11.41 &12.28 &6.00 &4.13 &17.18 &29.88 &8.80 &11.81 \\
TransCal~\cite{wang2020transferable} &4.95 &13.85 &16.58 &17.29 &17.34 &18.76 &18.77 &7.48 &19.54 &18.20 &7.13 &16.90 &14.73 \\
Ensemble~\cite{lakshminarayanan2017simple} &18.40 &7.47 &4.51 &7.82 &4.76 &4.24 &8.36 &17.96 &\textbf{3.92} &5.96 &17.68 &4.29 &8.78 \\
PseudoCal &\textbf{3.07} &\textbf{4.23} &5.28 &\textbf{1.96} &6.27 &5.70 &\textbf{2.52} &\textbf{4.05} &4.22 &\textbf{2.79} &\textbf{1.68} &7.03 &\textbf{4.07} \\
\cmidrule{0-13}
Oracle &2.38 &3.14 &2.34 &1.44 &1.92 &1.36 &1.98 &1.92 &1.37 &1.71 &1.43 &1.80 &1.90 \\
Accuracy &52.07 &74.48 &79.27 &64.24 &73.85 &75.42 &64.65 &50.65 &78.54 &70.37 &54.46 &81.48 &68.29 \\
\bottomrule[1pt]
\end{tabular}
$}
\end{table*}

\begin{table*}[!htbp]\centering
\caption{ECE ($\%$) of a closed-set UDA method BNM~\cite{cui2020towards} on \emph{Office-Home}.} 
\label{tab:full-uda-bnm-home}
\resizebox{1.0\textwidth}{!}{$
\begin{tabular}{l|rrrrrrrrrrrr|a}
\toprule[1pt]
Method &Ar $\to$ Cl &Ar $\to$ Pr &Ar $\to$ Re &Cl $\to$ Ar &Cl $\to$ Pr &Cl $\to$ Re &Pr $\to$ Ar &Pr $\to$ Cl &Pr $\to$ Re &Re $\to$ Ar &Re $\to$ Cl &Re $\to$ Pr & AVG \\
\cmidrule{0-13}
No Calib. &38.64 &22.49 &16.21 &30.89 &22.09 &18.25 &34.90 &42.46 &15.72 &27.11 &38.44 &14.52 &26.81 \\
MatrixScal~\cite{guo2017calibration} &39.37 &23.31 &19.01 &30.30 &25.73 &22.24 &31.37 &41.37 &15.98 &24.06 &37.39 &14.77 &27.07 \\
VectorScal~\cite{guo2017calibration} &30.83 &17.66 &9.97 &21.91 &16.40 &11.46 &27.76 &37.27 &12.36 &18.91 &29.06 &10.03 &20.30 \\
TempScal~\cite{guo2017calibration} &27.22 &16.34 &8.91 &20.39 &15.10 &10.21 &28.82 &35.60 &11.64 &20.12 &28.15 &9.67 &19.35 \\
CPCS~\cite{park2020calibrated} &33.80 &18.08 &8.12 &17.24 &19.77 &7.90 &28.68 &\textbf{17.28} &10.39 &28.36 &23.97 &6.86 &18.37 \\
TransCal~\cite{wang2020transferable} &25.75 &12.11 &5.87 &15.73 &10.51 &5.51 &21.41 &29.66 &5.02 &\textbf{15.17} &26.25 &4.80 &14.82 \\
Ensemble~\cite{lakshminarayanan2017simple} &29.52 &16.03 &12.00 &22.77 &15.55 &14.06 &25.17 &32.06 &11.53 &19.55 &30.46 &11.56 &20.02 \\
PseudoCal &\textbf{14.27} &\textbf{8.74} &\textbf{4.60} &\textbf{15.46} &\textbf{6.31} &\textbf{4.69} &\textbf{20.90} &18.35 &\textbf{4.76} &15.66 &\textbf{15.47} &\textbf{3.55} &\textbf{11.06} \\
\cmidrule{0-13}
Oracle &3.16 &2.18 &1.76 &2.00 &3.14 &1.95 &2.92 &1.78 &1.10 &1.68 &2.64 &1.77 &2.17 \\
Accuracy &54.39 &73.49 &79.78 &64.52 &73.69 &76.82 &61.68 &51.13 &80.35 &70.05 &55.56 &82.36 &68.65 \\
\bottomrule[1pt]
\end{tabular}
$}
\end{table*}

\begin{table*}[!htbp]\centering
\caption{ECE ($\%$) of a closed-set UDA method MCC~\cite{jin2020minimum} on \emph{Office-Home}.} 
\label{tab:full-uda-mcc-home}
\resizebox{1.0\textwidth}{!}{$
\begin{tabular}{l|rrrrrrrrrrrr|a}
\toprule[1pt]
Method &Ar $\to$ Cl &Ar $\to$ Pr &Ar $\to$ Re &Cl $\to$ Ar &Cl $\to$ Pr &Cl $\to$ Re &Pr $\to$ Ar &Pr $\to$ Cl &Pr $\to$ Re &Re $\to$ Ar &Re $\to$ Cl &Re $\to$ Pr & AVG \\
\cmidrule{0-13}
No Calib. &23.74 &14.31 &10.89 &12.70 &13.15 &11.72 &14.36 &23.18 &8.98 &12.69 &22.40 &9.54 &14.81 \\
MatrixScal~\cite{guo2017calibration} &37.39 &23.28 &19.95 &31.00 &27.75 &25.27 &26.13 &35.70 &16.27 &21.56 &35.20 &14.95 &26.20 \\
VectorScal~\cite{guo2017calibration} &21.05 &12.79 &7.87 &10.96 &11.18 &8.20 &16.87 &28.29 &9.64 &7.58 &21.40 &6.15 &13.50 \\
TempScal~\cite{guo2017calibration} &12.23 &6.43 &3.61 &4.06 &4.69 &\textbf{2.85} &11.38 &22.91 &5.83 &4.79 &13.60 &\textbf{4.11} &8.04 \\
CPCS~\cite{park2020calibrated} &25.11 &15.31 &\textbf{3.60} &19.41 &14.36 &4.49 &13.83 &35.66 &8.56 &24.08 &24.99 &14.27 &16.97 \\
TransCal~\cite{wang2020transferable} &3.04 &6.31 &5.98 &12.75 &7.42 &8.60 &11.95 &4.59 &9.90 &10.48 &3.95 &6.37 &7.61 \\
Ensemble~\cite{lakshminarayanan2017simple} &19.20 &11.30 &8.05 &10.01 &9.69 &8.51 &10.11 &18.98 &7.13 &9.15 &19.42 &7.44 &11.58 \\
PseudoCal &\textbf{2.71} &\textbf{5.04} &3.81 &\textbf{3.17} &\textbf{4.64} &3.06 &\textbf{2.66} &\textbf{1.54} &\textbf{3.85} &\textbf{2.73} &\textbf{2.51} &5.86 &\textbf{3.47} \\
\cmidrule{0-13}
Oracle &2.41 &2.57 &2.31 &2.67 &1.73 &1.62 &1.58 &0.84 &1.80 &2.51 &1.66 &2.35 &2.00 \\
Accuracy &47.26 &69.29 &75.90 &59.91 &68.33 &70.16 &56.32 &44.49 &76.04 &66.87 &50.65 &79.48 &63.73 \\
\bottomrule[1pt]
\end{tabular}
$}
\end{table*}

\begin{table*}[!htbp]\centering
\caption{ECE ($\%$) of a closed-set UDA method CDAN~\cite{long2018conditional} on \emph{Office-Home}.} 
\label{tab:full-uda-cdan-home}
\resizebox{1.0\textwidth}{!}{$
\begin{tabular}{l|rrrrrrrrrrrr|a}
\toprule[1pt]
Method &Ar $\to$ Cl &Ar $\to$ Pr &Ar $\to$ Re &Cl $\to$ Ar &Cl $\to$ Pr &Cl $\to$ Re &Pr $\to$ Ar &Pr $\to$ Cl &Pr $\to$ Re &Re $\to$ Ar &Re $\to$ Cl &Re $\to$ Pr & AVG \\
\cmidrule{0-13}
No Calib. &24.88 &14.66 &10.39 &14.71 &13.05 &11.25 &13.24 &22.54 &8.37 &12.19 &21.41 &8.74 &14.62  \\
MatrixScal~\cite{guo2017calibration} &35.03 &22.64 &19.14 &28.14 &26.14 &22.96 &24.20 &33.34 &15.03 &20.32 &30.69 &13.78 &24.28 \\
VectorScal~\cite{guo2017calibration} &18.81 &10.46 &7.24 &8.92 &9.81 &6.73 &15.31 &26.51 &9.18 &7.51 &16.70 &5.76 &11.91  \\
TempScal~\cite{guo2017calibration} &12.48 &5.82 &3.40 &\textbf{5.57} &5.14 &3.06 &9.78 &21.29 &6.12 &5.31 &12.55 &\textbf{4.06} &7.88 \\
CPCS~\cite{park2020calibrated} &31.45 &13.21 &2.36 &25.84 &24.68 &17.24 &13.44 &27.86 &10.09 &15.85 &41.38 &7.98 &19.28 \\
TransCal~\cite{wang2020transferable} &\textbf{2.65} &11.04 &11.67 &14.44 &13.41 &14.01 &16.34 &6.04 &15.50 &13.51 &5.46 &11.77 &11.32 \\
Ensemble~\cite{lakshminarayanan2017simple} &18.64 &11.85 &7.23 &10.87 &9.04 &7.94 &9.45 &19.12 &6.52 &9.90 &17.97 &6.56 &11.26 \\
PseudoCal &3.52 &\textbf{4.33} &\textbf{2.32} &5.67 &\textbf{4.81} &\textbf{2.82} &\textbf{6.36} &\textbf{3.78} &\textbf{2.05} &\textbf{3.28} &\textbf{3.85} &5.00 &\textbf{3.98} \\
\cmidrule{0-13}
Oracle &1.83 &2.96 &1.94 &3.88 &1.74 &2.20 &4.46 &3.22 &1.68 &2.50 &3.48 &2.08 &2.66 \\
Accuracy &48.00 &67.00 &75.07 &59.83 &66.88 &69.98 &58.59 &48.64 &76.31 &68.36 &53.33 &79.68 &64.31 \\
\bottomrule[1pt]
\end{tabular}
$}
\end{table*}

\begin{table*}[!htbp]\centering
\caption{ECE ($\%$) of a closed-set UDA method SAFN~\cite{xu2019larger} on \emph{Office-Home}.} 
\label{tab:full-uda-safn-home}
\resizebox{1.0\textwidth}{!}{$
\begin{tabular}{l|rrrrrrrrrrrr|a}
\toprule[1pt]
Method &Ar $\to$ Cl &Ar $\to$ Pr &Ar $\to$ Re &Cl $\to$ Ar &Cl $\to$ Pr &Cl $\to$ Re &Pr $\to$ Ar &Pr $\to$ Cl &Pr $\to$ Re &Re $\to$ Ar &Re $\to$ Cl &Re $\to$ Pr & AVG \\
\cmidrule{0-13}
No Calib. &28.25 &15.29 &12.40 &16.62 &14.10 &12.45 &18.17 &29.68 &10.94 &14.92 &25.77 &10.08 &17.39 \\
MatrixScal~\cite{guo2017calibration} &37.63 &23.66 &20.05 &28.07 &26.01 &23.00 &25.60 &37.84 &16.22 &20.98 &33.18 &14.69 &25.58 \\
VectorScal~\cite{guo2017calibration} &21.01 &12.78 &9.20 &10.96 &10.28 &7.67 &16.03 &26.93 &8.91 &10.72 &20.21 &6.35 &13.42  \\
TempScal~\cite{guo2017calibration} &12.33 &5.56 &3.17 &4.62 &4.22 &\textbf{3.40} &9.99 &21.72 &5.64 &6.36 &14.33 &3.89 &7.94 \\
CPCS~\cite{park2020calibrated} &31.45 &16.18 &10.90 &23.93 &11.19 &6.71 &15.78 &25.66 &18.73 &5.24 &34.50 &\textbf{2.80} &16.92 \\
TransCal~\cite{wang2020transferable} &7.50 &\textbf{4.23} &\textbf{2.80} &4.11 &\textbf{3.63} &4.89 &3.14 &7.47 &4.76 &3.26 &5.65 &3.46 &4.57 \\
Ensemble~\cite{lakshminarayanan2017simple} &25.00 &13.33 &9.91 &15.20 &11.62 &10.14 &16.12 &26.14 &9.54 &13.15 &23.56 &8.55 &15.19 \\
PseudoCal &\textbf{3.30} &6.41 &4.14 &\textbf{3.46} &7.06 &5.18 &\textbf{2.99} &\textbf{3.40} &\textbf{3.79} &\textbf{2.70} &\textbf{3.33} &7.12 &\textbf{4.41} \\
\cmidrule{0-13}
Oracle &3.10 &3.78 &1.94 &2.06 &1.85 &2.18 &2.65 &1.66 &1.11 &1.16 &2.68 &1.92 &2.17 \\
Accuracy &50.65 &70.96 &75.81 &64.44 &70.42 &72.30 &62.55 &49.55 &77.16 &70.54 &55.51 &79.97 &66.66 \\
\bottomrule[1pt]
\end{tabular}
$}
\end{table*}

\begin{table*}[!htbp]\centering
\caption{ECE ($\%$) of a closed-set UDA method MCD~\cite{saito2018maximum} on \emph{Office-Home}.} 
\label{tab:full-uda-mcd-home}
\resizebox{1.0\textwidth}{!}{$
\begin{tabular}{l|rrrrrrrrrrrr|a}
\toprule[1pt]
Method &Ar $\to$ Cl &Ar $\to$ Pr &Ar $\to$ Re &Cl $\to$ Ar &Cl $\to$ Pr &Cl $\to$ Re &Pr $\to$ Ar &Pr $\to$ Cl &Pr $\to$ Re &Re $\to$ Ar &Re $\to$ Cl &Re $\to$ Pr & AVG \\
\cmidrule{0-13}
No Calib. &26.24 &16.26 &12.30 &16.42 &14.19 &13.27 &19.02 &27.38 &10.35 &13.63 &24.25 &9.43 &16.89  \\
MatrixScal~\cite{guo2017calibration} &41.44 &28.57 &22.89 &34.21 &27.91 &26.19 &28.46 &39.91 &18.20 &22.91 &36.82 &16.58 &28.67  \\
VectorScal~\cite{guo2017calibration} &21.79 &12.62 &8.36 &11.89 &7.19 &7.75 &17.75 &27.43 &8.99 &10.10 &20.83 &5.72 &13.37 \\
TempScal~\cite{guo2017calibration} &8.59 &\textbf{4.59} &\textbf{2.87} &3.65 &\textbf{2.79} &\textbf{2.90} &10.42 &17.99 &4.85 &3.96 &9.86 &\textbf{3.29} &6.31 \\
CPCS~\cite{park2020calibrated} &20.66 &11.43 &21.72 &27.95 &11.22 &11.03 &24.03 &12.63 &10.13 &23.42 &48.48 &7.86 &19.21 \\
TransCal~\cite{wang2020transferable} &\textbf{2.43} &8.94 &9.45 &10.78 &10.81 &10.80 &9.86 &\textbf{2.07} &13.56 &11.69 &3.49 &11.19 &8.76 \\
Ensemble~\cite{lakshminarayanan2017simple} &20.49 &10.59 &7.24 &11.59 &9.53 &9.16 &15.53 &22.66 &6.52 &9.95 &19.45 &6.66 &12.45 \\
PseudoCal &2.52 &4.93 &3.93 &\textbf{3.39} &6.57 &3.70 &\textbf{5.05} &2.68 &\textbf{3.52} &\textbf{3.76} &\textbf{3.39} &7.28 &\textbf{4.23} \\
\cmidrule{0-13}
Oracle &2.22 &2.48 &2.08 &2.68 &2.31 &2.13 &3.02 &1.97 &2.44 &2.26 &2.61 &2.11 &2.36 \\
Accuracy &46.55 &63.75 &73.01 &57.44 &64.86 &67.45 &53.81 &42.77 &73.72 &65.88 &51.07 &77.63 &61.49 \\
\bottomrule[1pt]
\end{tabular}
$}
\end{table*}

\begin{table*}[!htbp]\centering
\caption{ECE ($\%$) of a closed-set UDA method ATDOC~\cite{liang2021domain} on \emph{DomainNet}.} 
\label{tab:full-uda-atdoc-dmnt}
\resizebox{0.63\textwidth}{!}{$
\begin{tabular}{l|ccccccc|a}
\toprule[1pt]
Method &C $\to$ S &P $\to$ C &P $\to$ R &R $\to$ C &R $\to$ P &R $\to$ S &S $\to$ P & AVG \\
\cmidrule{0-8}
No Calib. &12.22 &9.27 &3.75 &9.81 &6.85 &12.36 &7.92 &8.88  \\
MatrixScal~\cite{guo2017calibration} &34.30 &27.58 &15.58 &23.23 &18.37 &28.05 &27.44 &24.94  \\
VectorScal~\cite{guo2017calibration} &16.19 &11.45 &3.97 &15.11 &10.19 &19.26 &9.52 &12.24  \\
TempScal~\cite{guo2017calibration} &10.32 &6.52 &1.94 &10.86 &8.51 &13.31 &6.92 &8.34 \\
CPCS~\cite{park2020calibrated} &12.87 &13.31 &4.46 &8.25 &5.11 &13.90 &4.34 &8.89  \\
TransCal~\cite{wang2020transferable} &19.89 &23.51 &26.65 &22.52 &24.93 &19.46 &24.59 &23.08 \\
Ensemble~\cite{lakshminarayanan2017simple} &8.71 &5.73 &\textbf{1.59} &6.91 &4.41 &9.38 &4.66 &5.91 \\
PseudoCal &\textbf{1.68} &\textbf{1.98} &2.51 &\textbf{1.66} &\textbf{1.21} &\textbf{1.71} &\textbf{1.61} &\textbf{1.77} \\
\cmidrule{0-8}
Oracle &0.98 &1.92 &0.86 &1.18 &0.70 &1.16 &1.17 &1.14 \\
Accuracy &53.74 &56.51 &74.95 &55.59 &61.65 &50.41 &59.64 &58.93 \\
\bottomrule[1pt]
\end{tabular}
$}
\end{table*}

\begin{table*}[!htbp]\centering
\caption{ECE ($\%$) of a closed-set UDA method BNM~\cite{cui2020towards} on \emph{DomainNet}.} 
\label{tab:full-uda-bnm-dmnt}
\resizebox{0.63\textwidth}{!}{$
\begin{tabular}{l|rrrrrrr|a}
\toprule[1pt]
Method &C $\to$ S &P $\to$ C &P $\to$ R &R $\to$ C &R $\to$ P &R $\to$ S &S $\to$ P & AVG \\
\cmidrule{0-8}
No Calib. &30.88 &29.27 &15.37 &27.87 &21.79 &31.65 &22.41 &25.61 \\
MatrixScal~\cite{guo2017calibration} &37.91 &31.17 &18.31 &26.82 &22.33 &32.31 &28.64 &28.21 \\
VectorScal~\cite{guo2017calibration} &23.10 &20.02 &9.88 &21.80 &14.83 &26.68 &14.18 &18.64 \\
TempScal~\cite{guo2017calibration} &19.11 &18.79 &9.40 &19.28 &14.42 &21.49 &12.81 &16.47 \\
CPCS~\cite{park2020calibrated} &14.45 &13.75 &7.98 &\textbf{2.72} &4.35 &\textbf{4.14} &11.50 &8.41 \\
TransCal~\cite{wang2020transferable} &9.21 &\textbf{6.31} &\textbf{5.82} &6.73 &\textbf{1.69} &9.56 &\textbf{1.98} &\textbf{5.90} \\
Ensemble~\cite{lakshminarayanan2017simple} &25.08 &23.46 &12.61 &23.42 &18.52 &27.34 &18.70 &21.30\\
PseudoCal &\textbf{5.08} &12.43 &6.18 &8.10 &5.20 &6.64 &6.82 &7.21 \\
\cmidrule{0-8}
Oracle &1.60 &3.17 &3.40 &1.63 &1.50 &1.00 &1.81 &2.02 \\
Accuracy &52.90 &55.52 &74.30 &57.71 &63.95 &51.61 &62.30 &59.76 \\
\bottomrule[1pt]
\end{tabular}
$}
\end{table*}

\begin{table*}[!htbp]\centering
\caption{ECE ($\%$) of a closed-set UDA method MCC~\cite{jin2020minimum} on \emph{DomainNet}.} 
\label{tab:full-uda-mcc-dmnt}
\resizebox{0.63\textwidth}{!}{$
\begin{tabular}{l|rrrrrrr|a}
\toprule[1pt]
Method &C $\to$ S &P $\to$ C &P $\to$ R &R $\to$ C &R $\to$ P &R $\to$ S &S $\to$ P & AVG \\
\cmidrule{0-8}
No Calib. &15.19 &8.29 &4.79 &8.98 &6.91 &12.04 &8.63 &9.26  \\
MatrixScal~\cite{guo2017calibration} &36.95 &28.60 &15.99 &23.92 &18.95 &29.54 &28.72 &26.10  \\
VectorScal~\cite{guo2017calibration} &18.52 &11.63 &4.49 &15.98 &10.72 &20.86 &10.71 &13.27 \\
TempScal~\cite{guo2017calibration} &13.49 &5.92 &\textbf{2.36} &10.83 &8.96 &14.27 &7.67 &9.07 \\
CPCS~\cite{park2020calibrated} &29.26 &15.02 &3.44 &3.03 &6.00 &5.15 &2.66 &9.22 \\
TransCal~\cite{wang2020transferable} &16.89 &22.54 &23.45 &22.00 &24.68 &19.17 &23.44 &21.74\\
Ensemble~\cite{lakshminarayanan2017simple} &11.36 &5.38 &2.57 &6.03 &4.40 &9.32 &5.80 &6.41\\
PseudoCal &\textbf{2.72} &\textbf{1.45} &2.38 &\textbf{1.25} &\textbf{1.64} &\textbf{3.48} &\textbf{2.13} &\textbf{2.15} \\
\cmidrule{0-8}
Oracle &0.80 &1.36 &1.09 &0.96 &1.18 &0.97 &1.70 &1.15 \\
Accuracy &47.65 &51.27 &71.62 &50.51 &59.02 &45.14 &56.46 &54.52 \\
\bottomrule[1pt]
\end{tabular}
$}
\end{table*}

\begin{table*}[!htbp]\centering
\caption{ECE ($\%$) of a closed-set UDA method CDAN~\cite{long2018conditional} on \emph{DomainNet}.} 
\label{tab:full-uda-cdan-dmnt}
\resizebox{0.63\textwidth}{!}{$
\begin{tabular}{l|rrrrrrr|a}
\toprule[1pt]
Method &C $\to$ S &P $\to$ C &P $\to$ R &R $\to$ C &R $\to$ P &R $\to$ S &S $\to$ P & AVG \\
\cmidrule{0-8}
No Calib. &17.00 &10.51 &5.56 &9.83 &8.26 &11.88 &11.03 &10.58 \\
MatrixScal~\cite{guo2017calibration} &35.28 &27.82 &15.80 &22.11 &18.34 &27.24 &27.76 &24.91 \\
VectorScal~\cite{guo2017calibration} &17.44 &10.88 &4.37 &12.88 &9.45 &17.90 &9.81 &11.82  \\
TempScal~\cite{guo2017calibration} &13.39 &6.58 &2.75 &9.27 &8.30 &11.22 &8.32 &8.55  \\
CPCS~\cite{park2020calibrated} &\textbf{2.40} &17.27 &5.57 &4.24 &6.75 &11.42 &\textbf{1.81} &7.07 \\
TransCal~\cite{wang2020transferable} &14.85 &20.65 &22.93 &21.19 &22.27 &19.01 &20.55 &20.21 \\
Ensemble~\cite{lakshminarayanan2017simple} &12.96 &7.47 &3.54 &6.96 &5.73 &9.62 &7.75 &7.72 \\
PseudoCal &3.48 &\textbf{1.65} &\textbf{1.86} &\textbf{1.51} &\textbf{1.70} &\textbf{1.85} &2.08 &\textbf{2.02} \\
\cmidrule{0-8}
Oracle &1.03 &1.61 &1.07 &1.28 &0.73 &0.84 &1.43 &1.14 \\
Accuracy &49.07 &53.25 &71.82 &52.98 &60.75 &49.11 &57.51 &56.36 \\
\bottomrule[1pt]
\end{tabular}
$}
\end{table*}

\begin{table*}[!htbp]\centering
\caption{ECE ($\%$) of a closed-set UDA method SAFN~\cite{xu2019larger} on \emph{DomainNet}.} 
\label{tab:full-uda-safn-dmnt}
\resizebox{0.63\textwidth}{!}{$
\begin{tabular}{l|rrrrrrr|a}
\toprule[1pt]
Method &C $\to$ S &P $\to$ C &P $\to$ R &R $\to$ C &R $\to$ P &R $\to$ S &S $\to$ P & AVG \\
\cmidrule{0-8}
No Calib. &21.82 &17.98 &10.15 &17.90 &13.63 &20.70 &15.25 &16.78 \\
MatrixScal~\cite{guo2017calibration} &33.45 &22.54 &11.16 &21.05 &15.53 &26.33 &21.85 &21.70  \\
VectorScal~\cite{guo2017calibration} &19.61 &14.11 &4.73 &17.45 &10.40 &21.04 &10.49 &13.98 \\
TempScal~\cite{guo2017calibration} &15.12 &8.37 &4.12 &10.86 &8.23 &13.25 &8.07 &9.72 \\
CPCS~\cite{park2020calibrated} &21.96 &14.58 &8.22 &7.26 &7.52 &23.23 &4.31 &12.44 \\
TransCal~\cite{wang2020transferable} &6.58 &11.28 &14.28 &10.21 &12.67 &7.18 &13.10 &10.76 \\
Ensemble~\cite{lakshminarayanan2017simple} &19.74 &16.66 &9.08 &16.51 &12.48 &19.31 &14.03 &15.40 \\
PseudoCal &\textbf{3.40} &\textbf{4.44} &\textbf{1.50} &\textbf{2.23} &\textbf{0.81} &\textbf{2.12} &\textbf{1.79} &\textbf{2.33} \\
\cmidrule{0-8}
Oracle &0.86 &1.75 &1.21 &1.11 &0.78 &0.57 &1.06 &1.05  \\
Accuracy &48.14 &48.65 &66.40 &50.54 &59.89 &47.18 &56.17 &53.85 \\
\bottomrule[1pt]
\end{tabular}
$}
\end{table*}

\begin{table*}[!htbp]\centering
\caption{ECE ($\%$) of a closed-set UDA method MCD~\cite{saito2018maximum} on \emph{DomainNet}.} 
\label{tab:full-uda-mcd-dmnt}
\resizebox{0.63\textwidth}{!}{$
\begin{tabular}{l|rrrrrrr|a}
\toprule[1pt]
Method &C $\to$ S &P $\to$ C &P $\to$ R &R $\to$ C &R $\to$ P &R $\to$ S &S $\to$ P & AVG \\
\cmidrule{0-8}
No Calib. &12.97 &9.47 &3.80 &9.65 &7.01 &12.89 &7.80 &9.08 \\
MatrixScal~\cite{guo2017calibration} &31.47 &19.56 &10.05 &20.32 &14.30 &24.98 &18.45 &19.88  \\
VectorScal~\cite{guo2017calibration} &19.63 &12.59 &5.75 &16.53 &10.21 &20.95 &10.27 &13.70 \\
TempScal~\cite{guo2017calibration} &11.61 &5.39 &4.06 &7.58 &7.19 &10.79 &6.74 &7.62 \\
CPCS~\cite{park2020calibrated} &19.75 &6.09 &1.96 &7.94 &3.92 &23.82 &3.10 &9.51 \\
TransCal~\cite{wang2020transferable} &19.44 &21.53 &27.45 &21.44 &25.19 &18.45 &24.79 &22.61 \\
Ensemble~\cite{lakshminarayanan2017simple} &11.60 &7.54 &2.86 &6.95 &5.35 &11.07 &5.19 &7.22  \\
PseudoCal &\textbf{1.66} &\textbf{3.60} &\textbf{1.01} &\textbf{0.93} &\textbf{1.11} &\textbf{1.73} &\textbf{1.21} &\textbf{1.61} \\
\cmidrule{0-8}
Oracle &0.62 &1.81 &0.56 &0.85 &0.91 &0.73 &1.03 &0.93 \\
Accuracy &49.09 &48.21 &65.32 &49.49 &59.58 &46.81 &56.40 &53.56 \\
\bottomrule[1pt]
\end{tabular}
$}
\end{table*}

\begin{table*}[!htbp]\centering
\caption{ECE ($\%$) of closed-set UDA methods on \emph{Office-31}.} 
\label{tab:full-uda-office-0}
\resizebox{1.0\textwidth}{!}{$
\begin{tabular}{l|rrrra|rrrra|rrrra}
\toprule[1pt]
\multirow{2}{*}{Method} & \multicolumn{4}{c}{ATDOC~\cite{liang2021domain}} & & \multicolumn{4}{c}{BNM~\cite{cui2020towards}} & & \multicolumn{4}{c}{MCC~\cite{jin2020minimum}} & \\
&A $\to$ D &A $\to$ W &D $\to$ A &W $\to$ A &AVG &A $\to$ D &A $\to$ W &D $\to$ A &W $\to$ A &AVG &A $\to$ D &A $\to$ W &D $\to$ A &W $\to$ A &AVG \\
\cmidrule{0-15}
No Calib. &4.59 &6.66 &11.43 &12.91 &8.90 &11.12 &8.27 &24.60 &22.22 &16.55 &6.18 &7.80 &18.60 &19.97 &13.14 \\
MatrixScal~\cite{guo2017calibration} &9.58 &13.21 &14.04 &15.35 &13.05 &11.22 &8.81 &24.64 &21.94 &16.65 &9.70 &10.21 &18.99 &21.84 &15.19 \\
VectorScal~\cite{guo2017calibration} &4.57 &6.43 &15.69 &17.50 &11.05 &8.15 &4.11 &24.82 &23.59 &15.17 &5.12 &3.16 &20.53 &24.01 &13.21 \\
TempScal~\cite{guo2017calibration} &\textbf{3.39} &4.18 &24.37 &20.41 &13.09 &9.23 &4.98 &26.15 &21.55 &15.48 &3.79 &3.00 &22.07 &20.70 &12.39 \\
CPCS~\cite{park2020calibrated} &7.98 &8.94 &26.49 &22.80 &16.55 &11.65 &\textbf{2.02} &27.16 &17.73 &14.64 &4.69 &3.03 &29.84 &30.47 &17.01 \\
TransCal~\cite{wang2020transferable} &14.21 &14.64 &13.27 &11.02 &13.29 &\textbf{5.22} &2.70 &16.00 &13.72 &9.41 &3.77 &3.91 &5.57 &7.49 &5.19 \\
Ensemble~\cite{lakshminarayanan2017simple} &3.60 &\textbf{4.09} &9.04 &10.53 &6.82 &6.92 &4.63 &19.99 &19.56 &12.78 &3.07 &4.88 &17.18 &17.78 &10.73 \\
PseudoCal &6.64 &4.98 &\textbf{3.22} &\textbf{4.47} &\textbf{4.83} &6.30 &3.97 &\textbf{10.75} &\textbf{8.21} &\textbf{7.31} &\textbf{2.68} &\textbf{2.82} &\textbf{4.50} &\textbf{4.71} &\textbf{3.68} \\
\cmidrule{0-15}
Oracle &2.49 &3.15 &1.90 &2.35 &2.47 &2.65 &1.40 &2.63 &2.41 &2.27 &2.36 &2.67 &2.42 &2.05 &2.38 \\
Accuracy &91.57 &88.93 &73.41 &73.06 &81.74 &88.35 &90.94 &71.35 &73.77 &81.10 &91.37 &89.06 &69.86 &69.51 &79.95 \\
\bottomrule[1pt]
\end{tabular}
$}
\end{table*}

\begin{table*}[!htbp]\centering
\caption{ECE ($\%$) of closed-set UDA methods on \emph{Office-31}.} 
\label{tab:full-uda-office-1}
\resizebox{1.0\textwidth}{!}{$
\begin{tabular}{l|rrrra|rrrra|rrrra}
\toprule[1pt]
\multirow{2}{*}{Method} & \multicolumn{4}{c}{CDAN~\cite{long2018conditional}} & & \multicolumn{4}{c}{SAFN~\cite{xu2019larger}} & & \multicolumn{4}{c}{MCD~\cite{saito2018maximum}} & \\
&A $\to$ D &A $\to$ W &D $\to$ A &W $\to$ A &AVG &A $\to$ D &A $\to$ W &D $\to$ A &W $\to$ A &AVG &A $\to$ D &A $\to$ W &D $\to$ A &W $\to$ A &AVG \\
\cmidrule{0-15}
No Calib. &9.34 &7.96 &16.66 &17.39 &12.84 &6.17 &6.68 &20.34 &22.33 &13.88 &9.49 &8.88 &16.39 &17.03 &12.95  \\
MatrixScal~\cite{guo2017calibration} &11.90 &14.91 &17.21 &21.12 &16.29 &9.49 &13.97 &20.56 &23.43 &16.86 &9.83 &13.49 &17.86 &20.28 &15.37 \\
VectorScal~\cite{guo2017calibration} &6.04 &3.60 &17.67 &25.37 &13.17 &3.22 &\textbf{2.20} &21.07 &23.59 &12.52 &5.87 &4.61 &17.75 &20.52 &12.19  \\
TempScal~\cite{guo2017calibration} &5.70 &3.41 &16.10 &20.97 &11.55 &3.21 &2.83 &24.48 &23.41 &13.48 &\textbf{3.44} &\textbf{2.36} &32.09 &18.65 &14.14 \\
CPCS~\cite{park2020calibrated} &30.95 &5.67 &\textbf{4.99} &29.95 &17.89 &8.21 &18.21 &24.18 &22.12 &18.18 &11.85 &19.01 &32.45 &22.92 &21.56 \\
TransCal~\cite{wang2020transferable} &7.44 &6.84 &5.51 &\textbf{4.18} &5.99 &\textbf{3.04} &2.81 &6.43 &9.86 &5.54 &5.65 &4.76 &5.86 &4.39 &5.17 \\
Ensemble~\cite{lakshminarayanan2017simple} &4.98 &3.29 &7.41 &14.43 &7.53 &3.81 &5.75 &17.58 &20.20 &11.84 &6.25 &5.49 &13.53 &15.60 &10.22\\
PseudoCal &\textbf{4.78} &\textbf{3.04} &6.39 &6.78 &\textbf{5.25} &7.92 &5.51 &\textbf{4.00} &\textbf{4.26} &\textbf{5.42} &5.97 &5.33 &\textbf{4.38} &\textbf{4.06} &\textbf{4.94} \\
\cmidrule{0-15}
Oracle &3.26 &2.17 &2.94 &3.47 &2.96 &2.90 &1.75 &2.14 &2.27 &2.27 &3.55 &1.76 &2.31 &1.90 &2.38 \\
Accuracy &87.15 &87.17 &64.82 &67.23 &76.59 &89.96 &88.55 &69.33 &68.58 &79.11 &86.14 &85.53 &67.52 &66.63 &76.46 \\
\bottomrule[1pt]
\end{tabular}
$}
\end{table*}

\begin{table*}[!htbp]\centering
\caption{ECE ($\%$) of a partial-set UDA method ATDOC~\cite{liang2021domain} on \emph{Office-Home}.} 
\label{tab:full-pda-atdoc-home}
\resizebox{1.0\textwidth}{!}{$
\begin{tabular}{l|rrrrrrrrrrrr|a}
\toprule[1pt]
Method &Ar $\to$ Cl &Ar $\to$ Pr &Ar $\to$ Re &Cl $\to$ Ar &Cl $\to$ Pr &Cl $\to$ Re &Pr $\to$ Ar &Pr $\to$ Cl &Pr $\to$ Re &Re $\to$ Ar &Re $\to$ Cl &Re $\to$ Pr & AVG \\
\cmidrule{0-13}
No Calib. &28.21 &20.87 &10.76 &17.58 &23.49 &11.69 &19.16 &28.98 &14.34 &13.29 &28.22 &15.64 &19.35  \\
MatrixScal~\cite{guo2017calibration} &35.85 &19.37 &13.42 &29.69 &30.20 &21.94 &21.96 &37.00 &14.83 &19.36 &34.96 &16.94 &24.63 \\
VectorScal~\cite{guo2017calibration} &25.87 &15.83 &7.46 &18.37 &20.96 &11.63 &19.96 &33.03 &12.36 &11.16 &26.57 &11.61 &17.90  \\
TempScal~\cite{guo2017calibration} &21.08 &15.04 &5.75 &12.95 &17.86 &7.52 &18.23 &29.63 &12.88 &9.02 &23.66 &11.83 &15.45 \\
CPCS~\cite{park2020calibrated} &28.34 &27.40 &19.28 &14.37 &6.27 &10.86 &32.51 &39.04 &13.75 &11.28 &21.84 &7.92 &19.41 \\
TransCal~\cite{wang2020transferable} &\textbf{4.36} &\textbf{5.07} &10.58 &9.47 &\textbf{4.98} &12.82 &\textbf{9.12} &\textbf{5.81} &10.51 &13.32 &\textbf{5.34} &7.60 &8.25 \\
Ensemble~\cite{lakshminarayanan2017simple} &20.32 &12.06 &8.90 &11.80 &17.57 &7.89 &12.32 &22.25 &9.07 &11.81 &21.26 &10.68 &13.83  \\
PseudoCal &9.15 &7.08 &\textbf{3.21} &\textbf{7.59} &7.53 &\textbf{4.84} &11.80 &12.79 &\textbf{6.45} &\textbf{4.21} &10.75 &\textbf{4.10} &\textbf{7.46} \\
\cmidrule{0-13}
Oracle &3.09 &4.24 &2.82 &4.78 &4.93 &4.48 &4.04 &5.03 &4.94 &3.58 &5.24 &3.95 &4.26 \\
Accuracy &51.46 &64.99 &77.19 &61.89 &61.34 &73.44 &59.50 &49.01 &70.51 &67.68 &51.64 &71.43 &63.34 \\
\bottomrule[1pt]
\end{tabular}
$}
\end{table*}

\begin{table*}[!htbp]\centering
\caption{ECE ($\%$) of a partial-set UDA method MCC~\cite{jin2020minimum} on \emph{Office-Home}.} 
\label{tab:full-pda-mcc-home}
\resizebox{1.0\textwidth}{!}{$
\begin{tabular}{l|rrrrrrrrrrrr|a}
\toprule[1pt]
Method &Ar $\to$ Cl &Ar $\to$ Pr &Ar $\to$ Re &Cl $\to$ Ar &Cl $\to$ Pr &Cl $\to$ Re &Pr $\to$ Ar &Pr $\to$ Cl &Pr $\to$ Re &Re $\to$ Ar &Re $\to$ Cl &Re $\to$ Pr & AVG \\
\cmidrule{0-13}
No Calib. &22.91 &11.67 &8.45 &14.42 &14.34 &10.29 &12.63 &21.14 &8.22 &11.09 &22.46 &10.63 &14.02 \\
MatrixScal~\cite{guo2017calibration} &35.16 &19.13 &14.89 &29.94 &30.26 &25.30 &24.67 &34.81 &14.78 &18.58 &34.09 &15.73 &24.78 \\
VectorScal~\cite{guo2017calibration} &19.52 &9.73 &6.05 &12.79 &14.23 &11.07 &16.13 &26.53 &9.03 &9.29 &20.18 &7.95 &13.54  \\
TempScal~\cite{guo2017calibration} &13.14 &\textbf{5.37} &\textbf{3.05} &5.96 &6.62 &\textbf{4.21} &10.00 &20.08 &5.79 &5.39 &14.70 &6.12 &8.37 \\
CPCS~\cite{park2020calibrated} &19.34 &10.62 &4.00 &4.25 &\textbf{4.14} &12.00 &28.24 &37.75 &16.08 &5.70 &27.24 &12.51 &15.16 \\
TransCal~\cite{wang2020transferable} &2.74 &6.19 &5.25 &8.09 &5.92 &8.40 &11.03 &6.01 &7.29 &9.20 &\textbf{4.06} &\textbf{4.13} &6.53  \\
Ensemble~\cite{lakshminarayanan2017simple} &18.27 &9.86 &6.49 &9.68 &11.37 &7.27 &8.76 &18.05 &6.57 &9.21 &19.31 &9.10 &11.16 \\
PseudoCal &\textbf{2.51} &7.86 &4.70 &\textbf{3.04} &6.70 &5.78 &\textbf{4.20} &\textbf{4.01} &\textbf{3.96} &\textbf{3.99} &4.36 &6.23 &\textbf{4.78} \\
\cmidrule{0-13}
Oracle &2.29 &3.75 &2.04 &2.67 &3.07 &3.11 &2.69 &3.26 &1.97 &3.06 &3.47 &2.35 &2.81 \\
Accuracy &51.10 &74.17 &81.56 &62.53 &66.72 &73.16 &63.27 &50.03 &79.96 &70.80 &53.91 &79.33 &67.21 \\
\bottomrule[1pt]
\end{tabular}
$}
\end{table*}

\begin{table*}[!htbp]\centering
\caption{ECE ($\%$) of a partial-set UDA method PADA~\cite{cao2018partial} on \emph{Office-Home}.} 
\label{tab:full-pda-pada-home}
\resizebox{1.0\textwidth}{!}{$
\begin{tabular}{l|rrrrrrrrrrrr|a}
\toprule[1pt]
Method &Ar $\to$ Cl &Ar $\to$ Pr &Ar $\to$ Re &Cl $\to$ Ar &Cl $\to$ Pr &Cl $\to$ Re &Pr $\to$ Ar &Pr $\to$ Cl &Pr $\to$ Re &Re $\to$ Ar &Re $\to$ Cl &Re $\to$ Pr & AVG \\
\cmidrule{0-13}
No Calib. &20.35 &8.33 &5.30 &11.10 &12.28 &10.19 &8.93 &18.60 &4.83 &8.31 &18.33 &6.95 &11.13 \\
MatrixScal~\cite{guo2017calibration} &36.55 &24.04 &16.23 &34.97 &33.22 &28.87 &27.26 &37.58 &16.54 &20.45 &35.41 &16.45 &27.30  \\
VectorScal~\cite{guo2017calibration} &20.53 &7.22 &4.71 &12.28 &13.91 &13.44 &22.41 &31.95 &9.35 &9.07 &19.86 &8.57 &14.44  \\
TempScal~\cite{guo2017calibration} &15.15 &6.09 &3.34 &6.51 &6.43 &\textbf{4.64} &13.91 &23.77 &4.27 &6.34 &15.69 &6.11 &9.35 \\
CPCS~\cite{park2020calibrated} &24.22 &30.26 &24.81 &9.80 &7.37 &43.23 &28.84 &39.45 &14.97 &34.57 &4.55 &14.27 &23.03  \\
TransCal~\cite{wang2020transferable} &9.39 &23.43 &26.71 &21.37 &20.51 &21.88 &22.49 &11.25 &31.71 &24.23 &12.37 &25.06 &20.87  \\
Ensemble~\cite{lakshminarayanan2017simple} &11.42 &\textbf{4.97} &\textbf{2.88} &6.02 &\textbf{4.54} &4.65 &\textbf{3.76} &11.15 &\textbf{4.24} &6.13 &13.00 &\textbf{3.79} &\textbf{6.38} \\
PseudoCal &\textbf{2.95} &12.31 &7.51 &\textbf{4.68} &10.14 &5.38 &5.77 &\textbf{4.13} &7.19 &\textbf{3.71} &\textbf{3.28} &9.85 &6.41 \\
\cmidrule{0-13}
Oracle &2.16 &5.65 &2.27 &3.89 &5.70 &2.83 &5.06 &2.73 &3.98 &2.87 &3.06 &3.06 &3.61 \\
Accuracy &43.82 &59.83 &72.45 &51.70 &52.32 &58.14 &51.52 &40.66 &69.02 &63.73 &47.70 &71.54 &56.87  \\
\bottomrule[1pt]
\end{tabular}
$}
\end{table*}

\begin{table*}[!htbp]\centering
\caption{ECE ($\%$) of a white-box source-free UDA method SHOT~\cite{liang2020we} on \emph{DomainNet}.} 
\label{tab:full-sfda-shot-dmnt}
\resizebox{0.63\textwidth}{!}{$
\begin{tabular}{l|rrrrrrr|a}
\toprule[1pt]
Method &C $\to$ S &P $\to$ C &P $\to$ R &R $\to$ C &R $\to$ P &R $\to$ S &S $\to$ P & AVG \\
\cmidrule{0-8}
No Calib. &21.57 &16.14 &10.03 &18.18 &20.86 &24.71 &21.52 &19.00 \\
MatrixScal~\cite{guo2017calibration} &27.18 &19.67 &12.49 &19.13 &16.99 &21.60 &20.35 &19.63 \\
VectorScal~\cite{guo2017calibration} &17.79 &13.95 &6.46 &19.31 &16.25 &22.17 &13.20 &15.59  \\
TempScal~\cite{guo2017calibration} &13.91 &11.32 &4.81 &16.76 &16.47 &18.99 &10.63 &13.27 \\
CPCS~\cite{park2020calibrated} &12.52 &7.28 &4.93 &13.64 &10.86 &16.57 &9.10 &10.70\\
TransCal~\cite{wang2020transferable} &16.39 &23.80 &25.37 &24.23 &18.18 &15.87 &14.81 &19.81 \\
Ensemble~\cite{lakshminarayanan2017simple} &17.57 &13.24 &7.81 &15.24 &18.14 &21.40 &17.73 &15.88 \\
PseudoCal &\textbf{5.82} &\textbf{6.08} &\textbf{2.91} &\textbf{7.23} &\textbf{7.17} &\textbf{7.51} &\textbf{8.38} &\textbf{6.44} \\
\cmidrule{0-8}
Oracle &2.03 &3.69 &1.37 &2.85 &2.25 &2.33 &2.78 &2.47 \\
Accuracy &59.80 &66.79 &78.34 &66.25 &66.08 &59.48 &62.88 &65.66 \\
\bottomrule[1pt]
\end{tabular}
$}
\end{table*}

\begin{table*}[!htbp]\centering
\caption{ECE ($\%$) of a black-box source-free UDA method DINE~\cite{liang2022dine} on \emph{DomainNet}.} 
\label{tab:full-sfda-dine-dmnt}
\resizebox{0.63\textwidth}{!}{$
\begin{tabular}{l|rrrrrrr|a}
\toprule[1pt]
Method &C $\to$ S &P $\to$ C &P $\to$ R &R $\to$ C &R $\to$ P &R $\to$ S &S $\to$ P & AVG \\
\cmidrule{0-8}
No Calib. &31.91 &22.54 &12.39 &21.43 &20.63 &28.77 &24.38 &23.15 \\
Ensemble~\cite{lakshminarayanan2017simple} &26.38 &18.72 &10.83 &17.03 &17.53 &24.28 &20.18 &19.28 \\
PseudoCal &\textbf{17.86} &\textbf{15.12} &\textbf{5.30} &\textbf{13.71} &\textbf{11.14} &\textbf{14.44} &\textbf{14.75} &\textbf{13.19} \\
\cmidrule{0-8}
Oracle &1.35 &1.87 &1.29 &1.62 &1.94 &1.38 &1.65 &1.59 \\
Accuracy &54.26 &63.00 &80.69 &64.52 &67.13 &56.75 &63.81 &64.31  \\
\bottomrule[1pt]
\end{tabular}
$}
\end{table*}

\end{document}